\documentclass{article}

\usepackage{microtype}
\usepackage{graphicx}
\usepackage{subfigure}
\usepackage{booktabs} 

\usepackage{hyperref}

\usepackage[accepted]{icml2023}

\usepackage{amsmath}
\usepackage{amssymb}
\usepackage{mathtools}
\usepackage{amsthm}

\usepackage{url}
\usepackage{color}
\usepackage{makecell}
\usepackage{bbm}
\usepackage{amsmath}
\usepackage{arydshln}
\usepackage{multirow}
\usepackage{booktabs}
\usepackage{amssymb}
\usepackage{xcolor,pifont}
\usepackage{CJKutf8}
\usepackage[capitalize,noabbrev]{cleveref}

\theoremstyle{plain}
\newtheorem{theorem}{Theorem}

\theoremstyle{definition}

\theoremstyle{remark}

\usepackage[textsize=tiny]{todonotes}

\icmltitlerunning{Contrastive Preference Optimization: Pushing the Boundaries of LLM Performance in Machine Translation}

\newcommand*\colourcross[1]{%
  \expandafter\newcommand\csname #1cross\endcsname{\textcolor{#1}{\ding{56}}}%
}
\colourcross{red}
\newcommand*\colourcheck[1]{%
  \expandafter\newcommand\csname #1check\endcsname{\textcolor{#1}{\ding{52}}}%
}
\colourcheck{green}

\newcommand\extrafootertext[1]{%
    \bgroup
    \renewcommand\thefootnote{\fnsymbol{footnote}}%
    \renewcommand\thempfootnote{\fnsymbol{mpfootnote}}%
    \footnotetext[0]{#1}%
    \egroup
}

\definecolor{myDeepYellow}{rgb}{0.9412, 0.6902, 0.302}
\definecolor{myYellow}{rgb}{0.9765, 0.8824, 0.7255}
\definecolor{myShallowBlue}{HTML}{d3eaf2}
\definecolor{myBlue}{HTML}{92cbdf}

\newcommand\worse{\def\argi{}\docommandworse}
\def\docommandworse#1 {\colorbox{myYellow!75}{#1} \let\next\argi}
\def\argi{\let\next\docommandworse}

\def\docommandworst#1 {\colorbox{myYellow!75}{#1} \let\next\argii}
\def\argii{\let\next\docommandworst}

\newcommand\better{\def\argii{}\docommandbetter}
\def\docommandbetter#1 {\colorbox{myShallowBlue!80}{#1} \let\next\argii}
\def\argii{\let\next\docommandbetter}

\newcommand\best{\def\argii{}\docommandbest}
\def\docommandbest#1 {\colorbox{myBlue!100}{#1} \let\next\argii}
\def\argii{\let\next\docommandbest}

\newcommand{\compactworse}[1]{\setlength{\fboxsep}{1pt}\colorbox{myYellow! 75}{#1}}
\newcommand{\compactbetter}[1]{\setlength{\fboxsep}{1pt}\colorbox{myShallowBlue! 80}{#1}}
\newcommand{\compactbest}[1]{\setlength{\fboxsep}{1pt}\colorbox{myBlue! 100}{#1}}

\begin{document}

\twocolumn[
\icmltitle{Contrastive Preference Optimization: Pushing the Boundaries of LLM Performance in Machine Translation}



\icmlsetsymbol{equal}{*}

\begin{icmlauthorlist}
\icmlauthor{Haoran Xu$^{\spadesuit}$}{}
\icmlauthor{Amr Sharaf$^{\heartsuit}$}{}
\icmlauthor{Yunmo Chen$^{\spadesuit}$}{}
\icmlauthor{Weiting Tan$^{\spadesuit}$}{}
\icmlauthor{Lingfeng Shen$^{\spadesuit}$}{}
\icmlauthor{Benjamin Van Durme$^{\spadesuit}$}{}
\icmlauthor{Kenton Murray$^{*\ \spadesuit}$}{}
\icmlauthor{Young Jin Kim$^{*\ \heartsuit}$}{}
\end{icmlauthorlist}



\icmlcorrespondingauthor{Haoran Xu}{hxu64@jhu.edu}
\icmlcorrespondingauthor{Kenton Murray}{kenton@jhu.edu}
\icmlcorrespondingauthor{Young Jin Kim}{youki@microsoft.com}

\icmlkeywords{Machine Translation, Large Language Model}

\vskip 0.3in
]



\printAffiliationsAndNotice{\icmlEqualContribution$^{\spadesuit}$Johns Hopkins University $^{\heartsuit}$Microsoft. Work done during an internship at Microsoft} 

\begin{abstract}
Moderate-sized large language models (LLMs) -- those with 7B or 13B parameters -- exhibit promising machine translation (MT) performance. However, 
they do not match the performance of state-of-the-art conventional encoder-decoder translation models or larger-scale LLMs such as GPT-4 \citep{openai2023gpt4}. In this study, we bridge this performance gap. We first assess the shortcomings of supervised fine-tuning for LLMs in the MT task, emphasizing the quality issues present in the reference data, despite being human-generated. Then, in contrast to supervised fine-tuning which mimics reference translations, we introduce \textbf{Contrastive Preference Optimization (CPO)}, a novel approach that trains models to avoid generating adequate but not perfect translations. Applying CPO to ALMA \citep{alma} models with only 22K parallel sentences and tuning only 0.1\% parameters yields significant improvements. The resulting model, called \textbf{ALMA-R}, can match or exceed the performance of the WMT competition winners and GPT-4 on WMT'21, WMT'22 and WMT'23 test datasets.
\end{abstract}
\section{Introduction}

Machine translation (MT) predominantly utilizes transformer encoder-decoder architectures \citep{attention}, which is evident in prominent models such as NLLB-200 \citep{nllb}, M2M100 \citep{m2m}, BiBERT \citep{bibert}, and MT5 \citep{mt5}. However, the emergence of decoder-only large language models (LLMs) such as the GPT series \citep{gpt3_few_shot,openai2023gpt4}, Mistral \citep{mistral}, LLaMA series \citep{llama1,llama2}, Falcon \citep{falcon40b}, \textit{inter alia}, which have shown remarkable efficacy in various NLP tasks, which attracts the interest of developing machine translation with these decoder-only LLMs. Recent studies \citep{zhu2023multilingual,jiao2023chatgpt,gptmt,wmt23,wmt23_metric} indicate that larger LLMs such as GPT-3.5 (175B) and GPT-4 exhibit strong translation abilities. However, the performance of smaller-sized LLMs (7B or 13B) still falls short when compared to conventional translation models \citep{zhu2023multilingual}.

\begin{figure}[h]
    \centering
    \resizebox{1\linewidth}{!}{
    \includegraphics[width=7.5cm]{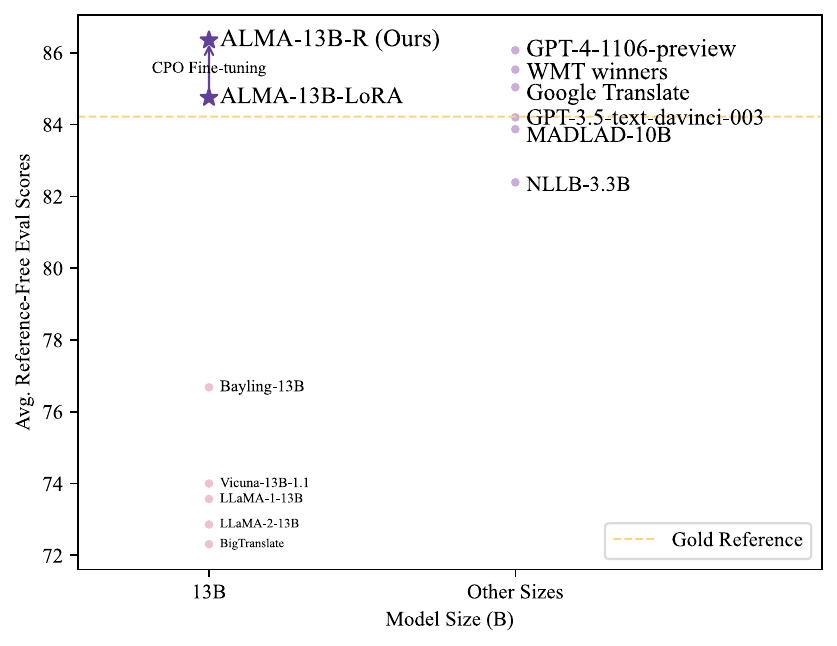}}
    \vskip -0.2in
    \caption{
    A performance comparison featuring our proposed model ALMA-13B-R against other recently released 13B LLM-based models, as well as top-performing translation systems like GPT-4 and WMT winners. This evaluation covers the WMT'22 test data across 8 directions, involving translations to and from English for German, Czech, Chinese, and Russian. Scores are averaged by three different reference-free models: \texttt{wmt23-cometkiwi-da-xxl}, \texttt{XCOMET-XXL}, and \texttt{wmt22-cometkiwi-da}, and are also averaged across all directions. 
    The gold reference is also evaluated due to the reference-free approach.
    Our model, ALMA-13B-R, developed by further training ALMA-13B-LoRA using our proposed CPO method, either matches or surpasses the most advanced translation models. We show the detailed numerical data for all systems presented in the figure in Appendix \ref{app:sec:full_results}.
    }
    \label{fig:intro}
\end{figure}

Therefore, there are studies intend to enhance the translation performance for these smaller LLMs \citep{bigtranslate,tim,swie,zhu2023extrapolating,li2023eliciting,parrot,bayling}, but their improvements are relatively modest, primarily due to the predominant pre-training of LLMs on English-centric datasets, resulting in limited linguistic diversity \citep{alma}. Addressing this limitation, \citet{alma} initially fine-tune LLaMA-2 \citep{llama2} with extensive non-English monolingual data to enhance their multilingual abilities, and then perform supervised fine-tune (SFT) with high-quality parallel data to instruct the model to generate translations. Their model, named ALMA, outperforms all prior moderated-size LLMs, and even larger models such as GPT-3.5, in the translation task. Nonetheless, the performance still lags behind leading translation models such as GPT-4 and WMT competition winners. Our study bridges this gap by further fine-tuning ALMA models with our novel training method \textbf{Contrastive Preference Optimization (CPO)} and minimal costs, i.e.,  only 12M learnable parameters (equivalent to 0.1\% of the original model size) and a 22K dataset for 10 directions. The fine-tuned model is referred to as \textbf{ALMA-R}. A detailed performance comparison is illustrated in Figure \ref{fig:intro}.

CPO aims to mitigate two fundamental shortcomings of SFT. First, SFT's methodology of minimizing the discrepancy between predicted outputs and gold-standard references inherently caps model performance at the quality level of the training data. This limitation is significant, as even human-written data, traditionally considered high-quality, is not immune to quality issues (more details in Section \ref{sec:gold_or_gilded}). For instance, one may notice that some strong translation models are capable of producing translations superior to the gold reference, as illustrated in Figure \ref{fig:intro}. Secondly, SFT lacks a mechanism to prevent the model from rejecting mistakes in translations. While strong translation models can produce high-quality translations, they occasionally exhibit minor errors, such as omitting parts of the translation. \textit{Preventing the production of these near-perfect but ultimately flawed translations is essential}. To overcome these issues, we introduce Contrastive Preference Optimization (CPO) to train the ALMA model using specially curated preference data. After CPO training, the ALMA-R model shows marked improvements, achieving performance levels that match or even surpass those of GPT-4 and WMT competition winners.

Our main contributions are summarized as follows:

\noindent\textbf{Are reference Gold or Gilded?} We conducted an in-depth analysis of the training data (FLORES-200 data) utilized by the ALMA model. We meticulously compared the quality of the reference translations with those generated by strong translation models. Our findings reveal that, in numerous instances, the quality of human-written parallel data is even inferior to that of system-generated translations. This observation underscores a critical insight: training models exclusively towards replicating reference translations may not be the most effective approach, and reliance on reference-based evaluation could be flawed.

\noindent\textbf{Pushing the Performance Boundary of SFT} We introduce Contrastive Preference Optimization, which offers advantages in terms of memory efficiency, speed, and, crucially, enhanced effectiveness in improving translation quality. CPO breaks the performance bottleneck inherent in SFT's reference-mimicking learning process and pushes the performance boundary of models that have reached saturation through SFT training.\footnote{We release our code and models at: \url{https://github.com/fe1ixxu/ALMA}.}

\noindent\textbf{Preference Data} We build and release a high-quality preference dataset for the machine translation area.

\section{Gold or Gilded? Scrutinizing Gold Reference Quality}
\label{sec:gold_or_gilded}
\begin{figure}[ht]
    \centering
    \resizebox{1\linewidth}{!}{
    \includegraphics[width=7.5cm]{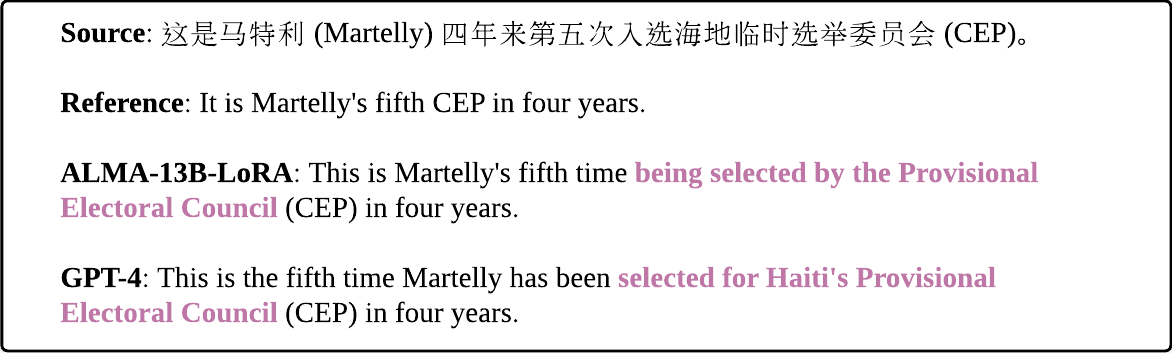}}
    \vskip -0.1in
    \caption{An example demonstrating that a human-written gold reference may not always be flawless, and could be surpassed by translations from advanced translation models. In this case, the reference retains the abbreviation ``CEP'' but fails to provide its full name. The highlighted phrases in the model-generated translations indicate the portions omitted by the gold reference.
    }
    \label{fig:example}
\end{figure}

The significance of target references is paramount in machine translation tasks. The paradigm of training models on the machine translation task heavily relies on the quality of the references since the model is commonly optimized using a loss that is defined to minimize the difference between the predicted outputs and gold reference. Consider a dataset $\mathcal{D}$, comprising pairs of source sentences $x$ and their corresponding target sentences (gold references) $y$, represented as  $\mathcal{D} = \left\{ x^{(i)}, y^{(i)} \right\}_{i=1}^{N}$, where $N$ is the total number of parallel sentences. The negative log-likelihood loss for these parallel sentences, in relation to a model $\pi_\theta$ parameterized by $\theta$, is defined as follows:
\begin{align}
 \mathcal{L}_{\text{NLL}} =  -\mathbb{E}_{(x,y) \sim \mathcal{D}} [\log \pi_\theta(y| x)].
\label{eq:task_def}
\end{align}
Hence, the ability of models to effectively translate is contingent upon the availability of high-quality translation pairs \citep{alma,maillard-etal-2023-small}. Furthermore, prevalent evaluation tools such as BLEU \citep{bleu} and COMET-22 \citep{comet22} predominantly rely on reference-based metrics. However, the precision of these evaluations is sensitive to and compromised by substandard references \citep{wmt23,wmt23_metric}. Recent research \citep{alma, wmt23,wmt23_metric} has shifted attention towards assessing the quality of parallel datasets, indicating that target references may not consistently represent the highest quality. In Figure \ref{fig:example}, we take a translation example from the FLORES-200 dataset, and compare the gold reference translation with outputs from the best ALMA model and GPT-4. This comparison reveals that the gold reference is a flawed translation, as it omits part of information, whereas the system-generated outputs demonstrate superior quality. This prompts an inquiry: \textit{Are references (even though human-written) truly equivalent to gold standards?} To thoroughly assess the quality of both the gold standard references and the outputs from contemporary high-performance translation models, we propose evaluating these outputs utilizing reference-free evaluation frameworks.

\paragraph{Models} We scrutinize the translation outputs from ALMA-13B-LoRA\footnote{ALMA-13B-LoRA is the best 13B translation model in the ALMA families. It initially undergoes \textit{full-weight} fine-tuning on monolingual data, followed by fine-tuning on high-quality human-written parallel data using \textit{low-rank adaptation} (LoRA) \citep{lora}.}, as well as zero-shot translations from the most recent GPT-4 (\texttt{gpt-4-1106-preview}). To assess the quality of these outputs, we employ two of the latest and largest reference-free models, each with a 10B parameter size and demonstrating very high correlation with human judgements \citep{wmt23_metric}. These models are \texttt{Unbabel/wmt23-cometkiwi-da-xxl} (henceforth referred to as \textbf{KIWI-XXL}) \citep{kiwi-xxl} and \texttt{Unbabel/XCOMET-XXL} (subsequently referred to as \textbf{XCOMET}) \citep{xcomet}.

\paragraph{Data} we consider the high-quality and human-written FLORES-200 dataset \citep{nllb}, comprising both development and test data, amounting to a total of 2009 samples for each language direction, to compare the gold references with the outputs generated by the models. We employed ALMA-13B-LoRA and GPT-4 to perform translations across five English-centric language pairs, covering both translations from and to English. These pairs include German (\texttt{de}), Czech (\texttt{cs}), Icelandic (\texttt{is}), Chinese (\texttt{zh}), and Russian (\texttt{ru}), with Icelandic (\texttt{is}) categorized as a low-resource language and the others as high-resource languages.

\paragraph{Prompt} The prompt employed for generating translations with ALMA models is consistent with the one used in \citet{alma}. For GPT-4 translation generation, we follow the guidelines suggested by \citet{gptmt}. The specifics of these prompts are detailed in Appendix \ref{app:sec:gpt_prompt}.

\paragraph{Model Outputs Can Be Better References}
In Table \ref{tab:gold_or_gilded}, we present the evaluation scores of KIWI-XXL and XCOMET for the gold references, ALMA-13B-LoRA outputs, and GPT-4 outputs. Additionally, we report \textit{Win Ratio}, reflecting the proportion of instances where model outputs surpass the gold standard references. These metrics are calculated as an average across five languages. Remarkably, even comparing with the high-quality Flores-200 dataset, the average performance of translation models in \texttt{xx}$\rightarrow$\texttt{en} translations significantly exceeds that of the references, showing approximately 3-4 point increases in KIWI-XXL and 4-6 point gains in XCOMET. Notably, a significant proportion of outputs are rated higher than the references by KIWI-XXL (e.g., \textbf{73.24\%} for ALMA), with a slightly reduced yet still substantial percentage when assessed using XCOMET (\textbf{60.17\%} for ALMA). In the \texttt{en}$\rightarrow$\texttt{xx} direction, while the overall performance between the translations from reference and two systems is comparable, approximately 40\% are still deemed superior to the reference translations.

\paragraph{Motivation: Help The Model Learn Rejection}
The aforementioned findings illustrate that translations produced by advanced models can sometimes surpass the quality of gold standard references. This raises the question of how to effectively utilize such data. A straightforward approach would involve fine-tuning the model using the source and the superior translations as references. While this could enhance the model's translation abilities, it does not equip the model with the discernment to identify and avoid generating suboptimal translations, exemplified by the ``good but not perfect" translations depicted in Figure \ref{fig:example}. Consequently, this situation motivates us to develop a new training objective, which aims to instruct the model in prioritizing the generation of higher-quality translations and rejecting lesser ones, in a style of contrastive learning with hard negative examples \citep{infoNCE,simclr,moco,hardexamples,tan2023multilingual}. This objective moves beyond the traditional focus on merely minimizing cross-entropy loss towards the reference.

\begin{table}[t]
\caption{
A performance comparison between gold references and outputs from advanced translation models, as assessed by two 10B-size reference-free evaluation models with the highest correlation to human preferences. The results indicate that the average performance of these strong translation models can even exceed that of the gold references, achieving a high success rate in beating the reference.
}
\vskip 0.15in
\label{tab:gold_or_gilded}
\centering

\resizebox{1\linewidth}{!}{
\begin{tabular}{lcccc}
\hline
              & KIWI-XXL & Win Ratio (\%) & XCOMET & \multicolumn{1}{l}{Win Ratio (\%)} \\
               \hline
\multicolumn{5}{c}{\textit{Translating to English} (\texttt{xx}$\rightarrow$\texttt{en})}                             \\
Reference     & 85.31    & -           & 88.82  & -                             \\
ALMA-13B-LoRA & 88.33    & 73.24       & 92.68  & 60.17                         \\
GPT-4         & 89.21    & 79.43       & 94.66  & 54.25                         \\
\hline
\multicolumn{5}{c}{\textit{Translating from English} (\texttt{en}$\rightarrow$\texttt{xx})}                           \\
Reference     & 87.85    & -           & 94.42  & -                             \\
ALMA-13B-LoRA & 85.62    & 42.15       & 93.07  & 35.46                         \\
GPT-4         & 87.30    & 49.13       & 94.21  & 38.09                         \\ \hline
\end{tabular}
}
\vskip -0.17in
\end{table}

\section{Contrastive Preference Optimization}
\label{sec:cpo}
To learn an objective that fosters superior translations and rejects inferior ones, access to labeled preference data is essential, yet such data is scarce in machine translation. In this section, we first describe the construction of our preference data and then introduces a preference learning technique, contrastive preference optimization (CPO).

\subsection{Triplet Preference Data}
\label{sec:build_pre_data}
We here detail our methodology for constructing preference data $\mathcal{D}$. This dataset is developed using the FLORES-200 data (both development and test sets) and encompasses the same language pairs as discussed in Section \ref{sec:gold_or_gilded}. For each language pair, the dataset comprises 2009 parallel sentences.

\begin{figure}[ht]
    \centering
    \resizebox{1\linewidth}{!}{
    \includegraphics[width=7.5cm]{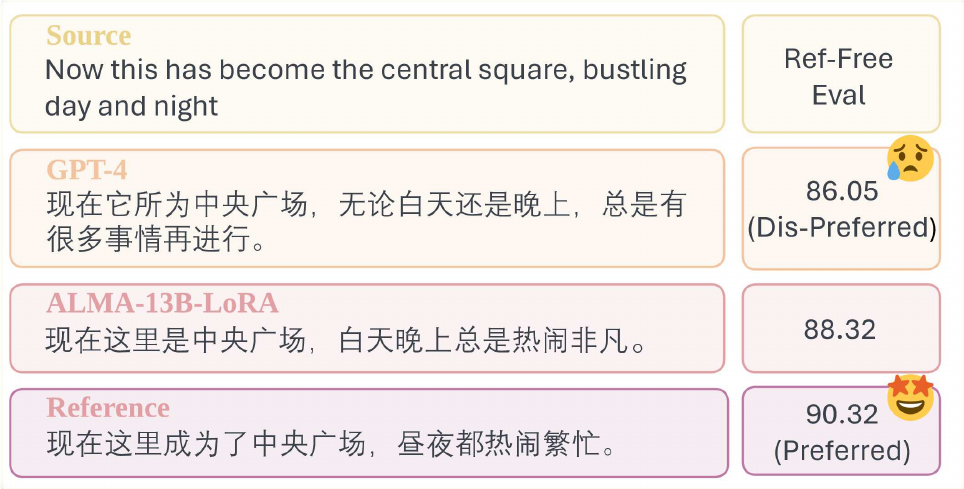}}
    \vskip -0.12in
    \caption{A triplet of translations, either model-generated or derived from a reference, accompanied by their respective scores as assessed by reference-free models. For a given source sentence, the translation with the highest score is designated as the preferred translation, while the one with the lowest score is considered dis-preferred, and the translation with a middle score is disregarded.
    }
    \label{fig:triplet}
\end{figure}

For a given source sentence $x$, whether translated from or to English, we utilize both GPT-4 and ALMA-13B-LoRA to generate respective translations, denoted as $y_\text{gpt-4}$ and $y_\text{alma}$. Together with the original target reference $y_\text{ref}$, this forms a triplet $\mathbf{y} = (y_\text{ref}, y_\text{gpt-4}, y_\text{alma})$, representing three different translation outputs for the input $x$. The reference-free evaluation models KIWI-XXL and XCOMET are then employed to score these translations, with the average scores represented as $\mathbf{s} = (s_\text{ref}, s_\text{gpt-4}, s_\text{alma})$.\footnote{The impact of using different evaluation models, such as only using XCOMET or KIWI-XXL, is explored in Section \ref{sec:metric_preferred}.} The highest-scoring translation is labeled as the preferred translation $y_w$, and the lowest-scoring as the dis-preferred translation $y_l$, i.e.,  $y_w = \mathbf{y}_{\arg\max_i(\mathbf{s})}, y_l = \mathbf{y}_{\arg\min_i(\mathbf{s})}$, where $i$ represents the index in the triplet. Translations with intermediate scores are not considered. An illustrative example of this selection process is depicted in Figure \ref{fig:triplet}. It is important to note that even the dis-preferred translations may be of high-quality. The designation 'dis-preferred' indicates that there is still room for improvement, perhaps through the addition of minor details. This approach of using high-quality but not flawless translations as dis-preferred data aids in training the model to refine details and achieve perfection in generated translations.

\subsection{Deriving the CPO Objective}
We discuss the derivation of CPO objective, beginning with an analysis of Direct Preference Optimization (DPO) \citep{dpo}. DPO represents a more direct optimization objective utilized in reinforcement learning from human feedback (RLHF) \citep{ziegler2019fine,ouyang2022training}. Given a set of source sentences $x$, alongside preferred translation targets $y_w$ and less preferred ones $y_l$, we can access a static dataset of comparisons, denoted as $\mathcal{D} = \left\{ x^{(i)}, y_{w}^{(i)}, y_{l}^{(i)} \right\}_{i=1}^{N}$. The loss function for DPO is constructed as a maximum likelihood objective for a parameterized policy $\pi_\theta$:

\begin{align}
\mathcal{L}(\pi_\theta;\pi_{\text{ref}}) = & -\mathbb{E}_{(x,y_w,y_l) \sim \mathcal{D}} \Big[ \log \sigma \Big( \beta \log \frac{\pi_{\theta}(y_w | x)}{\pi_{\text{ref}}(y_w | x)} \nonumber \\
& - \beta \log \frac{\pi_{\theta}(y_l | x)}{\pi_{\text{ref}} (y_l | x)} \Big) \Big],
\label{eq:dpo_loss}
\end{align}
where $\pi_{\text{ref}}$ is a pre-trained language (translation) model , $\sigma$ is the Sigmoid function, and $\beta$ is a hyperparameter. 
The DPO loss is derived a reparameterization process of the ground-truth reward and the corresponding optimal policy in the Proximal Policy Optimization (PPO) framework \citep{ppo}. As a result, 
DPO training can be conducted in a supervised fine-tuning style, as it relies exclusively on labeled preference data and does not require interaction between agents and their environment.

However, DPO has notable drawbacks compared to common SFT. Firstly, DPO is \textbf{memory-inefficient}: it necessitates twice the memory capacity to simultaneously store both the parameterized policy and the reference policy. Secondly, it is \textbf{speed-inefficient}: executing the model sequentially for two policies doubles the processing time. To address these inefficiencies, we introduce contrastive preference optimization.

The memory- or speed- inefficiency can be resolved when $\pi_\text{ref}$ is set as a uniform prior $U$, as the terms $\pi_{\text{ref}} (y_w | x)$ and $\pi_{\text{ref}} (y_l | x)$ cancel each other out. This negates the need for additional computations and storage beyond the policy model itself. Thus, we initially demonstrate that the DPO loss can be effectively approximated using a uniform reference model:
\begin{align}
\mathcal{L}(\pi_\theta;U) =  & -\mathbb{E}_{(x,y_w,y_l) \sim \mathcal{D}} \Big[ \log \sigma \Big( \beta \log \pi_{\theta}(y_w | x) \nonumber \\
& - \beta \log \pi_{\theta}(y_l | x) \Big) \Big].
\label{eq:dpo_uniform}
\end{align}

Specifically, we prove the below Theorem in Appendix \ref{app:simplify_dpo_loss}.

\begin{theorem}
When $\pi_\text{ref}$ is defined as $\pi_w$, an ideal policy that precisely aligns with the true data distribution of preferred data, the DPO loss $\mathcal{L}(\pi_\theta;\pi_w) + C$ is upper bounded by $\mathcal{L}(\pi_\theta;U)$, where $C$ is a constant.
\end{theorem}

The approximation in Equation \ref{eq:dpo_uniform} is effective because it minimizes the upper boundary of the DPO loss. The proof relies on an important assumption of  $\pi_\text{ref} = \pi_w$. Contrary to common practices where $\pi_\text{ref}$ is set as the initial SFT checkpoint, our approach considers it as the ideal policy we aim to reach. Although the ideal policy $\pi_w$ is unknown and unattainable during model training, it is not engaged in the loss after our approximation.

Furthermore, we incorporate a behavior cloning (BC) regularizer \citep{hejna2023contrastive} to ensure that $\pi_\theta$ does not deviate from the preferred data distribution:
\begin{align}
    & \min_\theta \mathcal{L}(\pi_\theta, U) \notag \\ & \text{  s.t.  } \mathbb{E}_{(x,y_w) \sim \mathcal{D}}\Big [ \mathbb{KL}(\pi_w(y_w|x)||\pi_\theta(y_w|x))\Big] < \epsilon,
\end{align}
where $\epsilon$ is a small positive constant and $\mathbb{KL}$ is Kullback–Leibler (KL) divergence. The regularizer can boil down to adding a SFT term on the preferred data (a detailed explanation is provided in Appendix \ref{app:simplify_dpo_loss}):
\begin{align}
    & \min_\theta\underbrace{ \mathcal{L}(\pi_\theta, U)}_{\mathcal{L}_\text{prefer}} \underbrace{-\mathbb{E}_{(x,y_w) \sim \mathcal{D}} [\log \pi_\theta(y_w| x)]}_{\mathcal{L}_\text{NLL}}.
\end{align}
The above is the formulation of our CPO loss, which includes one preference learning term $\mathcal{L}_{\text{prefer}}$ and one negative log likelihood term $\mathcal{L}_{\text{NLL}}$.

\section{Experiments}
\label{sec:experiments}
\subsection{Data}
Following Section \ref{sec:gold_or_gilded}, we consider 10 translation directions in the paper: \texttt{cs}$\leftrightarrow$\texttt{en}, \texttt{de}$\leftrightarrow$\texttt{en},
\texttt{is}$\leftrightarrow$\texttt{en},
\texttt{zh}$\leftrightarrow$\texttt{en},
\texttt{ru}$\leftrightarrow$\texttt{en}.
Building on the ALMA models' \citep{alma} insights that a small quantity of high-quality data can yield impressive translation results, our training dataset is even more compact. As detailed in Section \ref{sec:build_pre_data}, our preference training data is derived from the FLORES-200 dataset, a subset of which has been also employed in the training of ALMA models. This results in a total of $2\text{K}\times 10 \text{ directions} = 20\text{K}$ paired sentences. We detail the provenance distribution for each language pair from ALMA-13B-LoRA, GPT4, and reference as presented in Table \ref{tab:prefer_data_dis}. 
In addition to preference data assessed by large evaluation models, our dataset incorporates 1K internal human-labeled preference data, containing preferred and dis-preferred translations along with human preference. However, the human-labeled data is limited to just two translation directions: \texttt{en}$\rightarrow$\texttt{zh} and \texttt{en}$\rightarrow$\texttt{de}. 
The details regarding the composition and influence of human-labeled data are explored in Appendix \ref{app:sec:human_data}.\footnote{TL;DR: A brief overview of the impact of this human-labeled data suggests a minimal effect.}
In alignment with \citet{alma}, our primary focus is on the test set drawn from WMT'21 for \texttt{is} and WMT'22 for other languages. Additionally, we conduct auxiliary experiments evaluating models on WMT'23, covering six directions: \texttt{de}$\leftrightarrow$\texttt{en}, \texttt{zh}$\leftrightarrow$\texttt{en}, and \texttt{ru}$\leftrightarrow$\texttt{en}.

\begin{table}[t]
\caption{The provenance distribution for each language pair in the preference data.
}
\vskip 0.15in
\label{tab:prefer_data_dis}
\centering

\resizebox{0.88\linewidth}{!}{
\begin{tabular}{lccc}
\hline
                         & ALMA-13B-LoRA        & GPT-4       & Reference      \\ \hline
\texttt{en}$\leftrightarrow$\texttt{de} & 46\%  &  37\% &	17\%      \\
\texttt{en}$\leftrightarrow$\texttt{cs} & 32\%  &  41\% &	27\%       \\
\texttt{en}$\leftrightarrow$\texttt{is} & 36\%  &  40\% &	24\%        \\
\texttt{en}$\leftrightarrow$\texttt{zh} & 45\%  &  35\% &	20\%     \\
\texttt{en}$\leftrightarrow$\texttt{ru} & 31\%  &  44\% &	25\%         \\

\hline

\end{tabular}
}
\vskip -0.1in
\end{table}
\begin{table*}[t]
\caption{The overall results in \texttt{en}$\rightarrow$\texttt{xx} for WNT'21 and WMT'22. The application of the CPO method to fine-tune the ALMA-13B-LoRA model leads to a significant enhancement in performance, equalling or surpassing that of WMT competition winners and GPT-4. \textbf{Bold} numbers denote the highest scores across all systems. 
\compactbest{Dark blue boxes} indicates that the improvement over the original ALMA model achieves \textit{at least 80\% estimated accuracy} with the human judgement \citep{kocmi2024navigating}. 
Specifically, this denotes that for an agreement rate of 80\% with human decisions, the improvement needs a minimum of $\geq 1.24$ for both KIWI-XXL and XCOMET, and $\geq 0.53$ for KIWI-22. Further details on estimatied accuracy are provided in Appendix \ref{app:sec:estimated_acc}.
The lesser improvements are highlighted in \compactbetter{shallow blue boxes}.  Decreases in performance are marked with \compactworse{yellow boxes}.
}
\vskip 0.1in
\label{tab:main_en_xx}
 
\centering

\resizebox{1\linewidth}{!}{
\begin{tabular}{lccccccccc}
\hline
                         & \multicolumn{3}{c}{\texttt{de}}                                                                                                   & \multicolumn{3}{c}{\texttt{cs}}                                                                                                   & \multicolumn{3}{c}{\texttt{is}}                                                                                                   \\
\cmidrule(lr){2-4} \cmidrule(lr){5-7} \cmidrule(lr){8-10} 
\multirow{-2}{*}{Models} & KIWI-22                                & KIWI-XXL                               & XCOMET                                 & KIWI-22                                & KIWI-XXL                               & XCOMET                                 & KIWI-22                                & KIWI-XXL                               & XCOMET                                 \\ \hline
Gold Reference           & 82.67                                  & 84.01                                  & \textbf{97.85}                         & 83.19                                  & 81.83                                  & 90.27                                  & 80.51                                  & 85.20                                  & 91.52                                  \\
WMT Winners                 & \textbf{83.56}                         & 83.70                                  & 96.99                                  & 85.31                                  & \textbf{87.27}                         & \textbf{94.38}                         & 81.77                                  & 84.94                                  & 91.61                                  \\
GPT-4                    & 83.48                                  & \textbf{84.91}                                  & 97.56                                  & 84.81                                  & 85.35                                  & 93.48                                  & 81.03                                  & 81.21                                  & 90.00                                  \\
ALMA-13B-LoRA            & 82.62                                  & 81.64                                  & 96.49                                  & 84.14                                  & 84.24                                  & 92.38                                  & 81.71                                  & 83.31                                  & 91.20                                  \\ \hdashline
+ SFT on preferred data  & \better 82.75          & \better 81.85          & \better 96.67          & 84.14                                  & \worse 83.46          & \worse 91.99          & \worse 81.48          & \worse 82.11          & \worse 90.30          \\
+ DPO                    & \worse 82.40          & \worse 81.20          & \worse 96.40          & \worse 83.86          & \worse 83.45          & \worse 91.68          & \worse 81.43          & \worse 82.66          & \worse 90.33          \\
+ CPO (Ours, ALMA-13B-R)             & \best 83.28 & \best 84.25 & \better 97.48 & \best \textbf{84.99} & \best 87.06          & \best 93.61          & \better \textbf{82.18} & \best \textbf{85.68} & \better \textbf{91.93} \\ \hline
                         & \multicolumn{3}{c}{\texttt{zh}}                                                                                                   & \multicolumn{3}{c}{\texttt{ru}}                                                                                                   & \multicolumn{3}{c}{Avg.}                                                                                                 \\
\cmidrule(lr){2-4} \cmidrule(lr){5-7} \cmidrule(lr){8-10} 
\multirow{-2}{*}{Models} & KIWI-22                                & KIWI-XXL                               & XCOMET                                 & KIWI-22                                & KIWI-XXL                               & XCOMET                                 & KIWI-22                                & KIWI-XXL                               & XCOMET                                 \\ \hline
Gold Reference           & 80.92                                  & 81.70                                  & 90.42                                  & 82.96                                  & 84.62                                  & 94.17                                  & 82.05                                  & 83.47                                  & 92.85                                  \\
WMT Winners                 & 82.04                                  & 81.13                                  & 91.14                                  & \textbf{84.35}                         & 87.01                                  & 94.79                                  & \textbf{83.41}                         & 84.81                                  & 93.78                                  \\
GPT-4                    & 81.73                                  & 81.53                                  & 90.79                                  & 83.64                                  & 86.15                                  & 94.3                                   & 82.94                                  & 83.83                                  & 93.23                                  \\
ALMA-13B-LoRA            & 80.82                                  & 79.96                                  & 89.92                                  & 83.10                                  & 84.17                                  & 93.79                                  & 82.48                                  & 82.66                                  & 92.76                                  \\ \hdashline
+ SFT on preferred data  & \better 81.25          & \better 80.51          & \better 90.18          & \better 83.23          & \worse 84.15          & \worse 93.54          & \better 82.57          & \worse 82.42          & \worse 92.54          \\
+ DPO                    & \worse 80.74          & \worse 79.64          & \worse 89.58          & \worse 82.94          & \worse 83.40          & \worse 93.25          & \worse 82.27          & \worse 82.07          & \worse 92.25          \\
+ CPO (Ours, ALMA-13B-R)             & \best \textbf{82.25} & \best \textbf{84.32} & \best \textbf{92.03} & \best 83.98 & \best \textbf{87.37} & \best \textbf{95.22} & \best 83.34 & \best \textbf{85.74} & \best \textbf{94.05} \\ \hline
\end{tabular}
}
\vskip -0.1in
\end{table*}

\begin{table*}[t]
\caption{The overall results in \texttt{xx}$\rightarrow$\texttt{en} for WMT'21 and WMT'22. The usage of color and boldface are the same in Table \ref{tab:main_en_xx}.}
\vskip 0.1in
\label{tab:main_xx_en}
 
\centering

\resizebox{1\linewidth}{!}{
\begin{tabular}{lccccccccc}
\hline
\multirow{2}{*}{Models} & \multicolumn{3}{c}{\texttt{de}}      & \multicolumn{3}{c}{\texttt{cs}}      & \multicolumn{3}{c}{\texttt{is}}      \\
\cmidrule(lr){2-4} \cmidrule(lr){5-7} \cmidrule(lr){8-10} 
                        & KIWI-22 & KIWI-XXL & XCOMET & KIWI-22 & KIWI-XXL & XCOMET & KIWI-22 & KIWI-XXL & XCOMET \\ \hline
Gold Reference           & 78.74                                  & 78.56                         & 88.82                         & 82.08                                  & 83.11                                  & 84.60                                  & 80.88                                  & 85.04                         & 76.16                         \\
WMT Winners                 & 81.38                                  & 83.59                         & 93.74                         & 82.47                                  & 82.53                                  & 85.65                                  & 81.39                                  & 85.60                         & 78.14                         \\
GPT-4                    & \textbf{81.50}                         & \textbf{84.58}                & \textbf{94.47}                & 82.52                                  & 83.55                                  & \textbf{88.48}                         & 81.49                                  & \textbf{85.90}                & \textbf{81.11}                \\

ALMA-13B-LoRA            & 81.14                                  & 83.57                         & 93.30                         & 81.96                                  & 82.97                                  & 83.95                                  & 80.90                                  & 85.49                         & 76.68                         \\ \hdashline
+ SFT on preferred data  & \better 81.36          & \better 83.98 & \better 93.84 & \better 82.36          & \better 83.15          & \best 86.67 & \better 81.32          & \better 85.61 & \best 80.20 \\
+ DPO                    & \worse81.13          & \worse83.52 & \worse93.25 & \worse81.82          & \worse82.69          & \worse83.84          & \worse80.89          & \worse85.22 & \worse76.09 \\
+ CPO (Ours, ALMA-13B-R)             & \better \textbf{81.50} & \better 83.97 & \better 94.20 & \best \textbf{82.63} & \better \textbf{83.75} & \best 88.03          & \best \textbf{81.57} & \better 85.73 & \best 80.49 \\ \hline
                         & \multicolumn{3}{c}{\texttt{zh}}                                                                                 & \multicolumn{3}{c}{\texttt{ru}}                                                                                                   & \multicolumn{3}{c}{Avg.}                                                                               \\
                         \cmidrule(lr){2-4} \cmidrule(lr){5-7} \cmidrule(lr){8-10} 
\multirow{-2}{*}{Models} & KIWI-22                                & KIWI-XXL                      & XCOMET                        & KIWI-22                                & KIWI-XXL                               & XCOMET                                 & KIWI-22                                & KIWI-XXL                      & XCOMET                        \\ \hline
Gold Reference           & 77.09                                  & 74.19                         & 90.70                         & 80.74                                  & 79.59                                  & 88.56                                  & 79.91                                  & 80.10                         & 85.77                         \\
WMT Winners                 & 77.66                                  & 73.28                         & 87.2                          & 81.71                                  & 80.97                                  & 90.91                                  & 80.92                                  & 81.19                         & 87.13                         \\
GPT-4                    & \textbf{79.33}                         & \textbf{77.65}                & \textbf{92.06}                & 81.57                                  & 81.34                                  & 90.95                                  & 81.28                                  & \textbf{82.60}                & \textbf{89.41}                \\
ALMA-13B-LoRA            & 77.32                                  & 74.41                         & 89.88                         & 81.31                                  & 81.05                                  & 89.89                                  & 80.53                                  & 81.50                         & 86.74                         \\ \hdashline
+ SFT on preferred data  & \best 78.32 & \best 76.03 & \better 90.65 & \better 81.46          & \better 81.17          & \better 90.65          & \better 80.96          & \better 81.99 & \best 88.40 \\
+ DPO                    & \better 77.50          & \better 74.50 & \better 89.94 & \worse81.19          & \worse80.88          & \worse89.76          & \worse80.51          & \worse81.36 & \worse86.58 \\
+ CPO (Ours, ALMA-13B-R)            & \best 79.24          & \best 77.17 & \best 91.65 & \better \textbf{81.72} & \better \textbf{81.54} & \best \textbf{91.18} & \best \textbf{81.33} & \better 82.43 & \best 89.11 \\ \hline
\end{tabular}
}
\vskip -0.1in
\end{table*} 

\subsection{Training Setup}
We train the model in a \textit{many-to-many} multilingual machine translation manner, starting with ALMA-13B-LoRA as the initial checkpoint. During the training phase, we focus exclusively on updating the weights of the added LoRA parameters. These weights have a rank of 16 and only add an additional 12M parameters to the original 13B size of the model. We adhere to the default $\beta$ value of 0.1 as suggested by \citet{dpo}. The fine-tuning process of ALMA-13B-LoRA involves a batch size of 128, a warm-up ratio of 0.01, spanning a single epoch, and accommodating sequences with a maximum length of 512 tokens. To optimize training efficiency, we integrate the deepspeed tool \citep{deepspeed}. We utilize the same prompt as \citet{alma} and do not compute the loss for the prompt.  While our primary focus is on the performance of 13B models, CPO markedly benefits 7B models as well. Consequently, we also release ALMA-7B-R and provide a detailed discussion of its performance in Appendix \ref{app:sec:full_results}.

\subsection{Baselines}
\paragraph{SoTA Models} In this category, our benchmarks are established against, to the best of our knowledge, the strongest publicly available translation models. We first compare with \textbf{ALMA-13B-LoRA}, recognized as one of the top moderate-size language-model based translation systems, surpassing notable conventional models such as NLLB-54B in both WMT'21 and WMT'22. We also compare our results with \textbf{TowerInstruct}\footnote{\url{https://huggingface.co/datasets/Unbabel/TowerBlocks-v0.1}.}, a recently released LLM-based translation model and a contemporary work in the field.\footnote{ Note that TowerInstruct has used WMT'22 test data for training, so we exclude it from comparison on the WMT'22 test dataset.} Additionally, we evaluate against the zero-shot performance of the latest \textbf{GPT-4} (\texttt{gpt-4-1106-preview}), currently shown to be the best translation model among all LLM-based translation systems \citep{alma,bayling,tim,parrot}. Lastly, we include comparisons with the \textbf{WMT competition winners}, representing the highest standard of translation models within the competition, though it is noted that the winning models vary across different language directions.\footnote{The WMT winner systems used for comparison in each direction are provided in Appendix \ref{app:sec:wmt_winner}.}

\paragraph{SFT and DPO} We also compare different training objectives. Given that CPO is designed to steer learning towards preferred data, a straightforward benchmark is to compare its performance against directly SFT on the same preferred data set. Furthermore, considering that CPO is an evolution of DPO, we also include a comparative analysis with DPO.

\subsection{WMT'21 and WMT'22 Results}
We present the primary results for \texttt{en}$\rightarrow$\texttt{xx} and \texttt{xx}$\rightarrow$\texttt{en} in Table \ref{tab:main_en_xx} and Table \ref{tab:main_xx_en}, respectively. Our emphasis is primarily on reference-free evaluation models, due to our analysis in Section \ref{sec:gold_or_gilded}, which questions the reliability of gold references and highlights that evaluations can be compromised by poor-quality references \citep{wmt23,wmt23_metric}. However, we are not rejecting the use of reference-based models for evaluation but cautioning the potential pitfalls of poor-quality references. The reference-free models used for evaluation include KIWI-XXL, XCOMET, and a smaller yet popular model, \texttt{Unbabel/wmt22-cometkiwi-da} (hereinafter referred to as \textbf{KIWI-22}). Scores highlighted in \textbf{bold} represent the highest achieved across all systems. For a comprehensive comparison, we also include reference-based evaluations using sacreBLEU \citep{sacrebleu} and COMET-22 (\texttt{Unbabel/wmt22-comet-da}) \citep{comet22} in Appendix \ref{app:sec:full_results}.

\paragraph{Comparing With SoTA Models} While ALMA-13B-LoRA ranks as one of the top moderate-size LLM translation models, it slightly trails behind GPT-4 and the WMT competition winners. However, the incorporation of CPO significantly enhances ALMA's capabilities, bringing its performance to a level that is comparable to or even surpasses that of GPT-4 and WMT winners. For example, ALMA-13B-R achieves an average score of 85.74 on KIWI-XXL and 94.05 on XCOMET for \texttt{en}$\rightarrow$\texttt{xx} translations. These scores outperform GPT-4, which scores 83.83 on KIWI-XXL and 93.23 on XCOMET, as well as the WMT winners, who score 84.81 on KIWI-XXL and 93.78 on XCOMET.

\paragraph{Comparing With SFT and DPO} All training objectives in our study are fine-tuned using the ALMA-13B-LoRA model as a base. In Table \ref{tab:main_en_xx} and \ref{tab:main_xx_en}, 
we observe that SFT on preferred data marginally enhances the ALMA model's translation capability for \texttt{xx}$\rightarrow$\texttt{en}, and results in a slight deterioration for \texttt{en}$\rightarrow$\texttt{xx}. Similarly, DPO slightly decreases model performance. In contrast, CPO demonstrates significant improvements across all translation directions.

\subsection{WMT'23  Results}
We show the average results across all six directions in Table \ref{tab:wmt_23}, and provide the performance in each direction in Appendix \ref{app:sec:wmt23_results} due to the space constraint. Consistent with observations from WMT'21 and WMT'22, ALMA-13B-R surpasses contemporary moderate-size LLM-based translators such as ALMA-13B-LoRA and TowerInstruct, and either matches or exceeds WMT winners.

\begin{table}[t]
\caption{The average performance in WMT'23 across all 6 directions, with the highest score highlighted in bold.
}
\vskip 0.15in
\label{tab:wmt_23}
\centering

\resizebox{1\linewidth}{!}{
\begin{tabular}{lccc}
\hline
                         & KIWI-22        & KIWI-XXL       & XCOMET         \\ \hline
Gold Reference           & 78.74          & 75.56          & 86.30          \\
WMT Winners              & \textbf{80.57} & 77.72          & 88.24          \\
TowerInstruct            & 80.31          & 77.18          & 88.11          \\
ALMA-13B-LoRA            & 79.48          & 76.00          & 87.16          \\ \hdashline
+ CPO (Ours, ALMA-13B-R) & 80.55          & \textbf{78.97} & \textbf{89.74} \\ \hline
\end{tabular}
}
\vskip -0.1in
\end{table}

\section{Analyses}
All analyses use the WMT'21 and WMT'22 test sets, with their averaged performance being reported.
\subsection{Are Translations Really Better or Just Metric-Preferred?}
\label{sec:metric_preferred}
In our study, since the preferred data is selected by reference-free models and the same models are used for evaluation, we investigate the potential for ``cheating'' in the scoring process. Specifically, we question whether the translations become genuinely better or they simply align more closely with the evaluation model's preferences. This inquiry is addressed in two parts:

At the \underline{metric level}, we examine if training a model on data preferred by a specific metric (such as KIWI-XXL) yields improvements that are consistent across other metrics. To investigate this, we reconstruct the preference data using only KIWI-XXL or XCOMET and re-train the ALMA-13B-LoRA model using the CPO method. The results, presented in Table \ref{tab:metric_prefer}, do not indicate a significant bias towards the metric used for selecting preferred data. We observed similar and consistent improvements across all metrics, regardless of the specific metric used to select the preferred data. Considering Comet-series models may be positive correlated, we further evaluate ALMA-R using a non-comet metric, BLEURT \citep{bleurt}, and also observe significant improvements in Appendix \ref{app:sec:bleurt}. The inclusion of a third-party evaluation metric further substantiates the superior translation quality of ALMA-R.

At the \underline{method level}, we question whether training on metric-preferred data always leads to better scores on that metric, regardless of the method we use. Intriguingly,  we observe that generating translations favored by the metric — without true improvement — is not easy. For example, fine-tuning the model solely using DPO or SFT on metric-preferred data can even paradoxically lower its performance on this metric (in Table \ref{tab:main_en_xx}). This prompts us to question whether the improvements observed with CPO, an alternative objective that approximates DPO, when trained on the same data, are merely a result of metric bias. Our stance is that if both DPO and SFT fail to achieve improvements through metric bias, it stands to reason that CPO would similarly not benefit solely from such bias.

\begin{table}[t]
\caption{The influence of employing various reference-free models for creating preference data. The results illustrates that the final performance disparities are minimal whether using solely KIWI-XXL, XCOMET, or their combined ensemble.
}
\vskip 0.15in
\label{tab:metric_prefer}
\centering

\resizebox{1\linewidth}{!}{
\begin{tabular}{lccc}
\hline
Models for Building Preference Data & KIWI-22        & KIWI-XXL       & XCOMET         \\ \hline
\multicolumn{4}{c}{\textit{Translating to English} (\texttt{xx}$\rightarrow$\texttt{en})}                                    \\
N/A (ALMA-13B-LoRA baseline)        & 80.53          & 81.50          & 86.74          \\
KIWI-XXL                            & \textbf{81.33} & \textbf{82.59}          & 88.82          \\
XCOMET                              & 81.27          & 82.33          & \textbf{89.17} \\
Ensemble of above (Original)        & \textbf{81.33} & 82.43 & 89.11          \\ \hline
\multicolumn{4}{c}{\textit{Translating from English} (\texttt{en}$\rightarrow$\texttt{xx})}                                  \\
N/A (ALMA-13B-LoRA baseline)        & 82.48          & 82.66          & 92.76          \\
KIWI-XXL                            & 83.31          & \textbf{85.87}          & 93.97          \\
XCOMET                              & 83.09          & 85.43          & \textbf{94.09} \\
Ensemble of above (Original)        & \textbf{83.34} & 85.74 & 94.05          \\ \hline
\end{tabular}
}
\vskip -0.1in
\end{table}

\subsection{Human Evaluation}
The preceding analysis provides indirect evidence underscoring the absence of bias. Here, we incorporate human evaluation as direct proof.

we focused on the \texttt{zh}$\rightarrow$\texttt{en} direction, which aligns with the example presented in Section \ref{sec:gold_or_gilded}. We selected 400 samples from a total of 1875 test sentences, each sample including a Chinese source and two English translations, one from our base model ALMA-13B-LoRA and the other from ALMA-13B-R. Four bilingual (English and Chinese) speakers were enlisted to rate each translation on a scale from 0 to 6, as per the methodology outlined in \citet{kocmi-etal-2022-findings}.  We provide clarity on the evaluation criteria used for scoring in our study:
 
\noindent\textbf{0}: it signifies that the translation is nonsensical, failing to convey any coherent meaning.

\noindent\textbf{2}: it indicates that the translation partially preserves the meaning of the source text, albeit with substantial inaccuracies or omissions.

\noindent\textbf{4}: it denotes that the translation largely maintains the source text's meaning, with only minor issues such as slight grammatical errors.

\noindent\textbf{6}: it represents a perfect translation, accurately conveying the full meaning of the source text without any errors.

To ensure impartiality, each annotator was assigned 100 samples to score, with the order of the translations randomized to conceal their origin. In Table \ref{tab:human_eval}, we report the mean scores, rank position (with rank 1 indicating better translation and rank 2 indicating worse translation since we only have two translations to compare), and win ratio (note that both of them win if there is a tie) of ALMA and ALMA-R:

\begin{table}[t]
\caption{The results of human evaluation on sampled \texttt{zh}$\rightarrow$\texttt{en} WMT'22 test data. $\uparrow$ indicates that higher values are better, while $\downarrow$ indicates that lower values are better.
}
\vskip 0.15in
\label{tab:human_eval}
\centering

\resizebox{1\linewidth}{!}{
\begin{tabular}{lcccc}
\hline
                         & Avg. score $\uparrow$       & Avg. rank $\downarrow$      & Avg. win ratio (\%) & Ties (\%)         \\ \hline
ALMA-13B-LoRA &4.86 & 1.60&	62.50  &  40.30      \\
ALMA-13B-R   & \textbf{5.16} & \textbf{1.40} &  \textbf{77.80} & 40.30 \\\hline
\end{tabular}
}
\vskip -0.1in
\end{table}

The human evaluation results clearly demonstrate that ALMA-13-R outperforms the original ALMA-13B-LoRA. Consequently, our analysis supports the robustness and validity of using reference-free models like KIWI-XXL and XCOMET both for constructing preference data and for evaluation purposes.



\begin{figure*}[ht]
    \centering
    \resizebox{0.9\linewidth}{!}{
    \includegraphics[width=7.5cm]{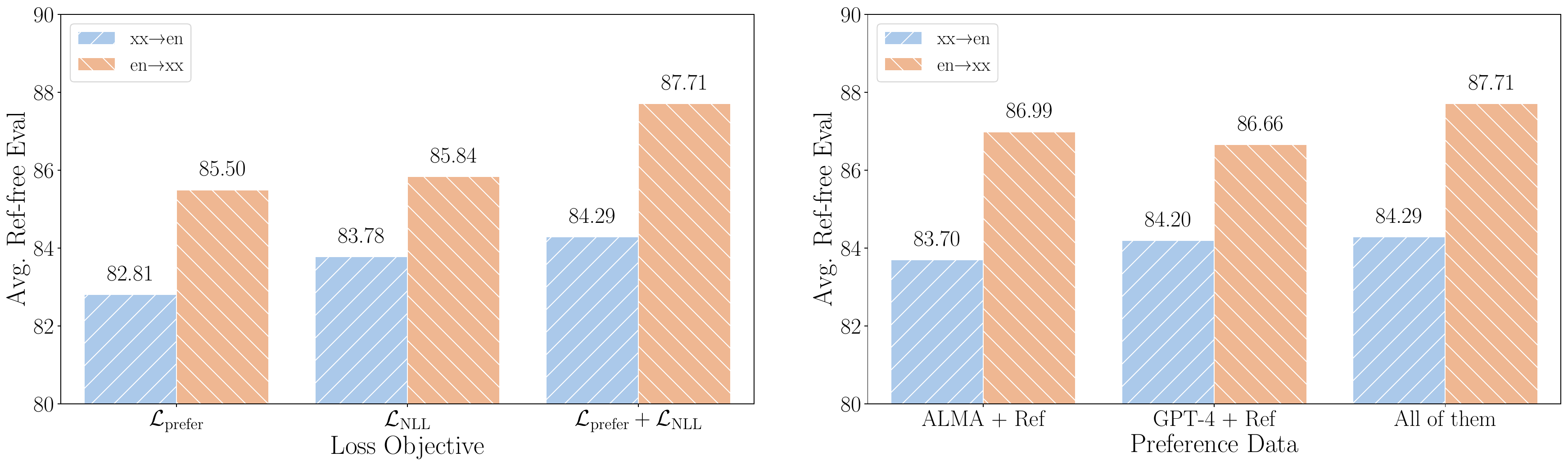}}
    \vskip -0.12in
    \caption{\textbf{Left:} an ablation study evaluating the significance of individual components in the CPO loss function, specifically analyzing how the preference learning loss $\mathcal{L}_\text{prefer}$ and the log-likelihood loss $\mathcal{L}_\text{NLL}$ each contribute to enhancing translation performance. \textbf{Right:} An ablation study assessing the significance of each component in the translation triplet. By excluding either ALMA or GPT-4 generated data from the preference triplet and re-training the model, we evaluate their respective impacts. The findings highlight the importance of ALMA-generated data for \texttt{en}$\rightarrow$\texttt{xx} translations and GPT-4 generated data for \texttt{xx}$\rightarrow$\texttt{en} translations.
    }
    \label{fig:ablation}
\end{figure*}

\subsection{Ablation Study}
\noindent\textbf{CPO Loss Components} 
The CPO loss function consists of two components: $\mathcal{L}_\text{prefer}$ for preference learning, and $\mathcal{L}_\text{NLL}$, which ensures the model does not deviate significantly from the preferred data distribution. To illustrate the significance of each term, we re-train the model exclusively with one of the components. It is important to note that training solely with $\mathcal{L}_\text{NLL}$ equates to the baseline scenario of SFT on preferred data. As depicted in the left of Figure \ref{fig:ablation}, the inclusion of both terms yields the optimal performance, while the absence of either leads to a decrease in performance. In Appendix \ref{app:sec:regularizer_dpo}, we also show that incorporating $\mathcal{L}_\text{NLL}$ into the DPO loss yields significant improvements.

\noindent\textbf{Preference Data Components}
Our preference data selection involves choosing preferred and dis-preferred translations from a triplet consisting of outputs from GPT-4, ALMA, and the gold reference. In the right of Figure \ref{fig:ablation}, we emphasize the significance of the data generated by both ALMA and GPT-4. The results indicate a notable decline in performance when ALMA data is excluded in the \texttt{en}$\rightarrow$\texttt{xx} direction.
Conversely, omitting GPT-4 data leads to a significant performance decrease in the \texttt{xx}$\rightarrow$\texttt{en} direction.
This demonstrates that data generated by both systems plays a helpful role in enhancing model performance.



\begin{table}[t]
\caption{An examination of the impact of dis-preferred data quality, contrasting noised data with natural, high-quality translations receiving the lowest scores as dis-preferred data. The findings underscore the importance of the quality of dis-preferred data.}
\vskip 0.15in
\label{tab:dis_preferred}
\centering

\resizebox{1\linewidth}{!}{
\begin{tabular}{lccc}
\hline
Dis-Preferred Data & KIWI-22        & KIWI-XXL       & XCOMET         \\ \hline
\multicolumn{4}{c}{\textit{Translating to English} (\texttt{xx}$\rightarrow$\texttt{en})}                   \\
Manually Noised    & 81.01          & 82.18          & 88.23          \\
Natural (Ours)     & \textbf{81.33} & \textbf{82.43} & \textbf{89.11} \\ \hline
\multicolumn{4}{c}{\textit{Translating from English} (\texttt{en}$\rightarrow$\texttt{xx})}                 \\
Manually Noised    & 82.71          & 83.13          & 92.80          \\
Natural (Ours)     & \textbf{83.34} & \textbf{85.74} & \textbf{94.05} \\ \hline
\end{tabular}
}
\vskip -0.1in
\end{table}

\subsection{Does The Quality of Dis-preferred Data Matter?}
In our experimental setup, dis-preferred data, though originating from strong translation models, receives the lowest scores when compared with two other translation outputs. A pertinent question arises: does the quality of dis-preferred data significantly impact model performance, and can high-quality (albeit imperfect) dis-preferred data aid in translation improvement? To explore this, we constructed a new set of preference data where the dis-preferred translations ($y_l$) are artificially generated, as opposed to being naturally derived high-quality translations.

In this new dataset, the preferred translation ($y_w$) remains the best of the three translation candidates, selected in the same manner as in Section \ref{sec:build_pre_data}. However, the dis-preferred translation is intentionally modified to be a noised version of $y_w$. We applied random deletions of words with a probability of 0.15 and word swaps within a range of 1 with a probability of 0.3, following the method suggested by \citet{tim} for creating manually noised dis-preferred data. This approach produces worse translations that are artificial.

Table \ref{tab:dis_preferred} compares the performance when using these manually noised dis-preferred data versus the original, naturally occurring high-quality dis-preferred data. The results show a substantial decline in performance across all three metrics and both translation directions when the dis-preferred data is manually noised, underscoring the importance of the quality of dis-preferred data in enhancing translation performance.

\section{Conclusion}
In this study, we initially proposed the potential quality issues of gold references in the MT task, highlighting instances where advanced translation models can outperform these references. This finding not only challenges model training via SFT,
but also the evaluation procedure that uses reference-based metrics. Subsequently, we introduce Contrastive Preference Optimization, a more efficient variant of of DPO. This method leverages both model-generated and reference data to guide the model in avoiding near-perfect yet flawed translations and learning superior ones. Our developed model, ALMA-13B-R, stands out as the first moderate-size LLM-based translation model to match, and in some cases surpass, the performance of GPT-4 and WMT competition winners, marking a significant advancement in the field of MT.

\section*{Impact Statement}
This paper presents work whose goal is to advance the field of Machine Translation and Large Language Model. There are many potential societal consequences of our work, none which we feel must be specifically highlighted here.

\section*{Acknowledgements}

We express our profound appreciation to annoymous reviewers for their helpful suggestions. We also thank Tianjian Li, Hieu Hoang, Marcin Junczys-Dowmunt, Huda Khayrallah, Thamme Gowda, Vikas Raunak, Matt Post, Anoop Kunchukuttan, Roman Grundkiewicz, Philipp Koehn, Hany Hassan Awadalla, Arul Menezes, and Vishal Chowdhary for their engaging and valuable discussions that greatly enriched our work. Special thanks to Tom Kocmi for his innovative suggestion to enhance numerical data visibility using a dynamic threshold determined by estimated accuracy. Our gratitude also extends to Pushpendre Rastogi and Joey Hejna for their insightful recommendations on the CPO theory. Furthermore, we acknowledge the Unbabel Team for their valuable advice on incorporating non-COMET metrics into our analysis.


\bibliography{example_paper}
\bibliographystyle{icml2023}

\newpage
\appendix
\onecolumn
\section*{Appendix Contents}
\appendix
\begin{table}[ht]
    \centering
    \footnotesize
    \begin{tabular}{cl}
    \textbf{Appendix Sections}    & \textbf{Contents}  \\ \toprule
    \autoref{app:sec:full_results} &  \begin{tabular}[c]{@{}l@{}} Comprehensive Results of WMT'21 and WMT'22 \end{tabular} \\ \midrule
    \autoref{app:sec:gpt_prompt} &  \begin{tabular}[c]{@{}l@{}} Prompts Used for Translations \end{tabular} \\ \midrule
     \autoref{app:simplify_dpo_loss} &  \begin{tabular}[c]{@{}l@{}} Theory of The CPO Loss  \end{tabular} \\ \midrule
    \autoref{app:sec:human_data}     &  \begin{tabular}[c]{@{}l@{}} Details and Influence of Human-Labeled Data\end{tabular} \\ \midrule
     \autoref{app:sec:wmt_winner}     &  \begin{tabular}[c]{@{}l@{}} Information of WMT winners\end{tabular} \\ \midrule
     \autoref{app:sec:estimated_acc}     &  \begin{tabular}[c]{@{}l@{}} Estimated Accuracy with Human Agreements\end{tabular} \\ \midrule
    \autoref{app:sec:wmt23_results}     &  \begin{tabular}[c]{@{}l@{}} Experimental Results on WMT'23\end{tabular} \\ \midrule
     \autoref{app:sec:bleurt}     &  \begin{tabular}[c]{@{}l@{}} Evaluation on Non-Comet Metrics\end{tabular} \\ \midrule
     \autoref{app:sec:regularizer_dpo}     &  \begin{tabular}[c]{@{}l@{}} Effectiveness of The BC Regularizer for DPO\end{tabular} \\ 
    \bottomrule
\end{tabular}    
\end{table}

\section{Comprehensive Results of WMT'21 and WMT'22}
\label{app:sec:full_results}
We show the comprehensive results for \texttt{en}$\rightarrow$\texttt{xx} In Table \ref{app:tab:full_en_xx} and \texttt{xx}$\rightarrow$\texttt{en} in Table \ref{app:tab:full_xx_en}. In this section, our study additionally includes results from recently released LLM-based translators, including Bayling-13B \citep{bayling}, BigTranslate \citep{bigtranslate}, ALMA-13B-LoRA \citep{alma}, the zero-shot performances of LLaMA-1-13B \citep{llama1} and LLaMA-2-13B \citep{llama2}. We also compare these with the most advanced current translation models, such as WMT competition winners, GPT-4, GPT-3.5-\texttt{text-davinci-003}, Google Translate, NLLB-3.3B, and MADLAD-10B \citep{madlad}. Importantly, we also present the performance of \textbf{ALMA-7B-R} here, which is fine-tuning on AMLA-7B-LoRA with CPO method. Except for reference-free evaluation, we also report two commonly used reference-based metrics, sacreBLEU \cite{sacrebleu,bleu} and COMET-22 \citep{comet22}.

\noindent\textbf{Introducing ALMA-7B-R}
In this study, we extend the ALMA-13B-R training methodology to a 7B model size, specifically fine-tuning ALMA-7B-LoRA using the CPO method with the same preference data as ALMA-13B-R. Consistent with our findings from ALMA-13B-R, the application of CPO significantly enhances performance. 

\noindent\textbf{Comparing with Advanced Translation Models}
Our model, ALMA-13B-R, is benchmarked against the most advanced current models, demonstrating performance comparable to GPT-4 and WMT winners. It surpasses leading commercial translation tools such as Google Translate in many cases and top multilingual translation models like NLLB, MADLAD-10B and GPT-3.5.

\noindent\textbf{Stop Using BLEU}
BLEU, a metric extensively utilized for decades, often diverges from neural-based and reference-free metrics, a phenomenon also observed in previous studies \citep{alma,wmt23_metric}. For instance, WMT competition winners often exhibit superior performance according to BLEU (or COMET-22), yet this is not corroborated by reference-free models. A case in point is the WMT winners scoring an exceptionally high 64.14 BLEU in \texttt{cs}$\rightarrow$\texttt{en} translations, significantly outperforming other models by 20 BLEU points. However, reference-free evaluations suggest these translations are inferior to those generated by our models and GPT-4. We hypothesize that this discrepancy may arise from WMT models being trained on domain-specific data closely related to the WMT test set, leading to high lexical matches but lacking semantic depth as evaluated by neural-based metrics. While BLEU scores are effective for assessing basic functionality in weaker models, their utility diminishes with advanced translation models capable of generating diverse translations. In such contexts, relying solely on BLEU for evaluation appears increasingly outdated.

\noindent\textbf{Towards Reference-Free Metrics}
Neural-based, reference-dependent metrics like COMET-22 demonstrate greater consistency with reference-free metrics and robustness compared to BLEU. For instance, with COMET-22, our models show significant improvements like other reference-free models over ALMA-13B-LoRA and comparable performance to GPT-4, e.g., 87.74 (Ours) vs. 87.68 (GPT-4) when \texttt{en}$\rightarrow$\texttt{xx}. However, it is important to note that, according to reference-free metrics, gold references are often inferior to system-generated translations, potentially indicating quality issues in the references that could impact COMET-22 evaluations. Consequently, inconsistencies still exist between COMET-22 and reference-free models like XCOMET. For example, XCOMET rates ALMA-R model on average higher than WMT winners (89.11 vs. 87.13), while COMET-22 favors WMT winners (85.21 vs. 85.60). In line with the recommendations in \citet{wmt23_metric}, we advocate for the use of reference-free models to circumvent the potential quality issues of references.
\newpage
\begin{table}[H]
\caption{The full results in \texttt{en}$\rightarrow$\texttt{xx} for WMT'21 and WMT'22 including both reference-free and reference-based metrics. \textbf{Bold} numbers denote the highest scores across all systems. \compactbest{Dark blue boxes} indicates that the improvement over the original ALMA model achieves \textit{at least 80\% estimated accuracy} with the human judgement \citep{kocmi2024navigating}, while the lesser improvements are highlighted in \compactbetter{shallow blue boxes}.  Decreases in performance are marked with \compactworse{yellow boxes}. The asterisk (*) indicates that we directly utilized the reported translation outputs from \citet{bayling} for evaluation purposes. Consequently, some baseline results for the \texttt{is} language are omitted in these instances. }
\label{app:tab:full_en_xx}
\vskip 0.04in
\centering

\resizebox{1\linewidth}{!}{

\begin{tabular}{lcccccccccc}
\hline
                        & \multicolumn{5}{c}{\texttt{de}}                                                                                                                                                                                                                          & \multicolumn{5}{c}{\texttt{cs}}                                                                                                                                                                                                                                               \\
                        \cmidrule(lr){2-6} \cmidrule(lr){7-11} 
                        & \multicolumn{1}{l}{BLEU}      & \multicolumn{1}{l}{COMET-22}           & KIWI-22                                                & KIWI-XXL                                             & XCOMET                                                 & \multicolumn{1}{l}{BLEU}                             & \multicolumn{1}{l}{COMET-22}           & KIWI-22                                                & KIWI-XXL                                             & XCOMET                                               \\ \hline
Gold Reference          & -                             & -                                      & 82.67                                                  & 84.01                                                & \textbf{97.85}                        & -                                                    & -                                      & 83.19                                                  & 81.83                                                & 90.27                                                \\
\hline

\hline
\end{tabular}

}

\end{table}

\begin{table}[H]
\caption{The full results in \texttt{xx}$\rightarrow$\texttt{en} for WMT'21 and WMT'22 including both reference-free and reference-based metrics. The usage of color and boldface are the same in Table \ref{app:tab:full_en_xx}. The asterisk (*) indicates that we directly utilized the reported translation outputs from \citet{bayling} for evaluation purposes. Consequently, some baseline results for the \texttt{is} language are omitted in these instances. }
\vskip 0.15in
\label{app:tab:full_xx_en}
\centering

\resizebox{1\linewidth}{!}{
\begin{tabular}{lcccccccccc}
\hline
                        & \multicolumn{5}{c}{\texttt{de}}                                                                                                                                                                                                                          & \multicolumn{5}{c}{\texttt{cs}}                                                                                                                                                                                            \\
                         \cmidrule(lr){2-6} \cmidrule(lr){7-11} 
                        & \multicolumn{1}{l}{BLEU}      & \multicolumn{1}{l}{COMET-22}           & KIWI-22                                                & KIWI-XXL                                               & XCOMET                                               & \multicolumn{1}{l}{BLEU}      & \multicolumn{1}{l}{COMET-22}  & KIWI-22                                                & KIWI-XXL                                               & XCOMET                          \\ \hline
Gold Reference          & -                             & -                                      & 78.74                                                  & 78.56                                                  & 88.82                                                & -                             & -                             & 82.08                                                  & 83.11                                                  & 84.60                           \\
\hline
\end{tabular}
}
\vskip -0.1in
\end{table}
\newpage
\section{Prompts for Translations}
\label{app:sec:gpt_prompt}
\begin{figure}[ht]
    \centering
    \resizebox{0.7\linewidth}{!}{
    \includegraphics[width=7.5cm]{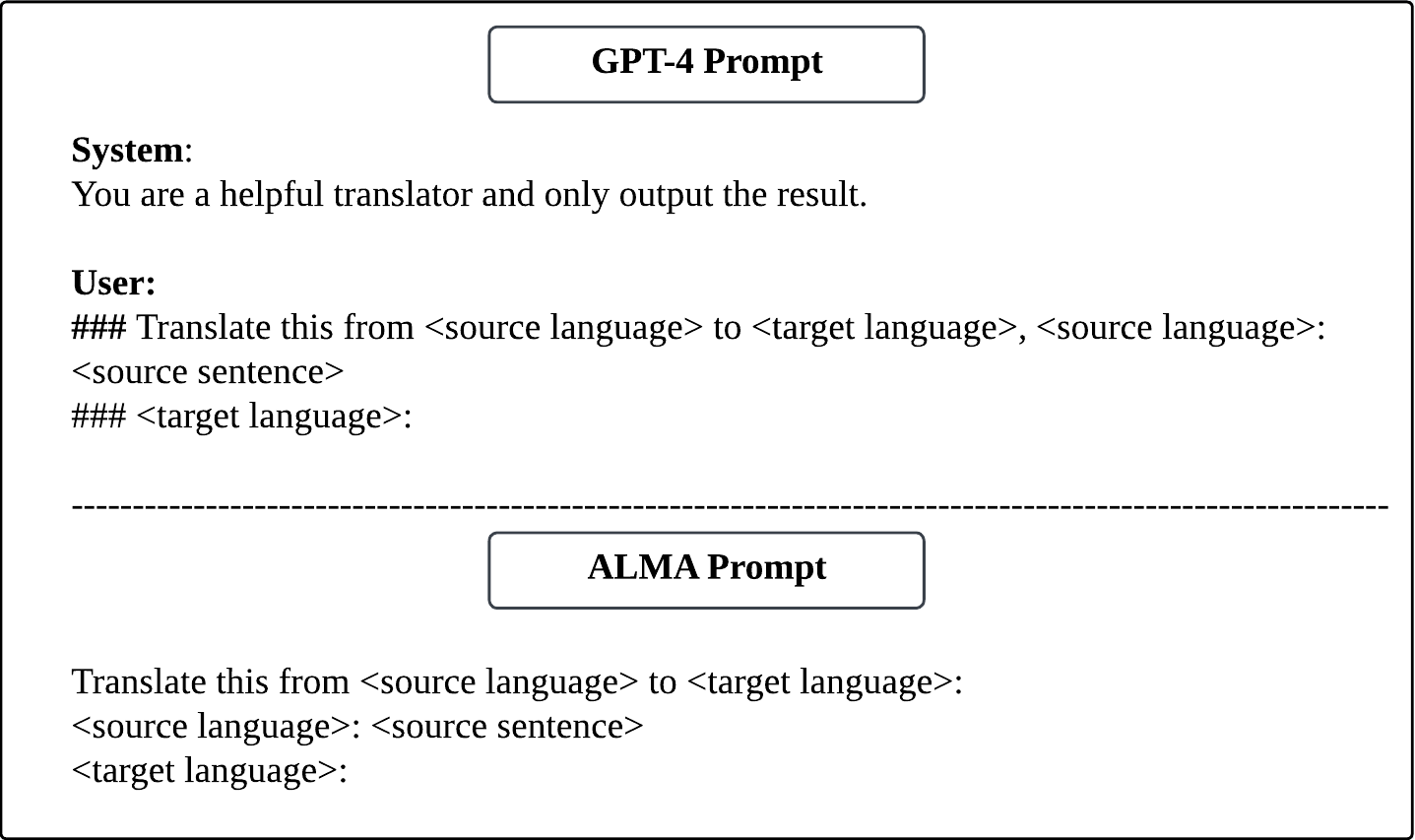}}
    \caption{The prompts employed for GPT-4 and ALMA models to perform translations.
    }
    \label{app:fig:gpt_prompt}
\end{figure}

Adhering to the prompt format for translation as utilized by \citet{gptmt} for GPT models, we employ the same prompt for GPT-4 in our study. Similarly, we use the same prompt employed by \citet{alma} for ALMA models. Prompts are depicted in Figure \ref{app:fig:gpt_prompt}.

\section{Theory}
\label{app:simplify_dpo_loss}
\subsection{Proof of The Upper Boundary}

\setcounter{theorem}{0} 
\begin{theorem}
When $\pi_\text{ref}$ is defined as $\pi_w$, an ideal policy that precisely aligns with the true data distribution of preferred data, the DPO loss $\mathcal{L}(\pi_\theta;\pi_w) + C$ is upper bounded by $\mathcal{L}(\pi_\theta;U)$, where $C$ is a constant.
\end{theorem}
\textit{Proof.}  $\pi_w$ represents an ideal policy that perfectly aligns the true data distribution of the preferred data. Hence, for any given data point $(x, y_w, y_l)$ from the preference dataset $\mathcal{D}$, the conditions $\pi_w(y_w | x)=1$ and $0\leq \pi_w(y_l | x) \leq 1$ hold true. Consequently, under this setup, the predictions for preferred data do not require reweighting by the reference model, and the DPO loss $\mathcal{L}(\pi_\theta;\pi_w)$ can be reformulated as follows :
\begin{align*}
\mathcal{L}(\pi_\theta;\pi_w) & = -\mathbb{E}_{(x,y_w,y_l) \sim \mathcal{D}} \Big[ \log \sigma \Big( \beta \log \frac{\pi_{\theta}(y_w | x)}{\pi_w(y_w | x)} \nonumber - \beta \log \frac{\pi_{\theta}(y_l | x)}{\pi_w (y_l | x)} \Big) \Big] \\
& = -\mathbb{E}_{(x,y_w,y_l) \sim \mathcal{D}} \Big[ \log \sigma \Big( \beta \log \pi_{\theta}(y_w | x) \nonumber - \beta \log \pi_{\theta}(y_l | x) + \beta\log\pi_w(y_l|x) \Big) \Big]. 
\end{align*}

After expanding the Sigmoid function, the loss becomes to:
\begin{align*}
    \mathcal{L}(\pi_\theta;\pi_w) & =  -\mathbb{E}_{(x,y_w,y_l) \sim \mathcal{D}} \Big[ \log\Big(\frac{1}{1+e^{-\beta \log \pi_{\theta}(y_w | x) \nonumber + \beta \log \pi_{\theta}(y_l | x) - \beta\log\pi_w(y_l|x)}}\Big )  \Big] \\
      & = -\mathbb{E}_{(x,y_w,y_l) \sim \mathcal{D}} \Big[ \log \Big ( \frac{1}{1 + \frac{\pi_\theta(y_l|x)^\beta}{\pi_\theta(y_w|x)^\beta \cdot \pi_w(y_l|x)^\beta}} \Big ) \Big] \notag\\
       & = -\mathbb{E}_{(x,y_w,y_l) \sim \mathcal{D}} \Big[ \log\pi_\theta(y_w|x)^\beta + \log\pi_w(y_l|x)^\beta - \log\Big(\pi_\theta(y_w|x)^\beta\cdot\pi_w(y_l|x)^\beta + \pi_\theta(y_l|x)^\beta \Big)\Big].
\end{align*}

Given that $\pi_w$ is a fixed model and $\log\pi_w(y_l|x)^\beta$ does not participate in gradient calculations or parameter updates, the above loss function is equivalent when we omit the term $\log\pi_w(y_l|x)^\beta$. Therefore, optimizing $\mathcal{L}(\pi_\theta;\pi_w)$ is equivalent to optimizing $\mathcal{L'}(\pi_\theta;\pi_w)$ as we define below:
\begin{align*}
    \mathcal{L'}(\pi_\theta;\pi_w) &\stackrel{\Delta}{=} \mathcal{L}(\pi_\theta;\pi_w) + \underbrace{\mathbb{E}_{(x,y_l) \sim \mathcal{D}} \Big[ \log\pi_w(y_l|x)^\beta  \Big ]}_{\text{$C$ in the Theorem}}\\
    & = -\mathbb{E}_{(x,y_w,y_l) \sim \mathcal{D}} \Big[ \log\pi_\theta(y_w|x)^\beta - \log\Big(\pi_\theta(y_w|x)^\beta\cdot\pi_w(y_l|x)^\beta + \pi_\theta(y_l|x)^\beta \Big)\Big].
\end{align*}
Considering that $0 \leq\pi_w(y_l|x) \leq 1$, the loss can be upper bounded as follows:
\begin{align*}
\label{app:eq:final}
    \mathcal{L'}(\pi_\theta;\pi_w) & \leq -\mathbb{E}_{(x,y_w,y_l) \sim \mathcal{D}} \Big[ \log\pi_\theta(y_w|x)^\beta - \log\Big(\pi_\theta(y_w|x)^\beta\cdot 1 + \pi_\theta(y_l|x)^\beta \Big)\Big] \notag\\
    & = -\mathbb{E}_{(x,y_w,y_l) \sim \mathcal{D}} \Big[ \log \sigma \Big( \beta \log \pi_{\theta}(y_w | x) \nonumber - \beta \log \pi_{\theta}(y_l | x) \Big) \Big] \\
    & = \mathcal{L}(\pi_\theta;U).
\end{align*}

Therefore, $\mathcal{L}(\pi_\theta;\pi_w) + C$ is upper bounded by $\mathcal{L}(\pi_\theta;U)$, where $C=\mathbb{E}_{(x,y_l) \sim \mathcal{D}} \Big[ \log\pi_w(y_l|x)^\beta  \Big ]$.

\subsection{BC Regularizer Simplification}
The contrastive preference optimization is originally defined as minimizing $\mathcal{L}(\pi_\theta;U)$ under the constraint of minimizing the difference between preferred data distribution and outputs of the learnable policy:
\begin{align*}
    \min_\theta \mathcal{L}(\pi_\theta, U)  \text{  s.t.  } \mathbb{E}_{(x,y_w) \sim \mathcal{D}}\Big [ \mathbb{KL}(\pi_w(y_w|x)||\pi_\theta(y_w|x))\Big] < \epsilon.
\end{align*}
This is equivalent to the following objective via Lagrangian duality:
\begin{align*}
    \min_\theta \mathcal{L}(\pi_\theta, U) + \lambda\cdot\mathbb{E}_{(x,y_w) \sim \mathcal{D}}\Big [ \mathbb{KL}(\pi_w(y_w|x)||\pi_\theta(y_w|x))\Big],
\end{align*}
where $\lambda$ is a hyperparamter and we set to 1. The optimization can be further optimized by expanding the KL divergence:
\begin{align*}
    \mathcal{L}_{\text{CPO}} & = \mathcal{L}(\pi_\theta, U) + \mathbb{E}_{(x,y_w) \sim \mathcal{D}}\Big [ \mathbb{KL}(\pi_w(y_w|x)||\pi_\theta(y_w|x))\Big] \\
    & =  \mathcal{L}(\pi_\theta, U) + \mathbb{E}_{(x,y_w) \sim \mathcal{D}}\Big [ \pi_w(y_w|x)\cdot\log\Big(\pi_w(y_w|x)\Big) -  \pi_w(y_w|x)\cdot\log\Big(\pi_\theta(y_w|x)\Big) \Big] \\
    & =  \mathcal{L}(\pi_\theta, U) + \mathbb{E}_{(x,y_w) \sim \mathcal{D}}\Big [ 1\cdot0-1\cdot\log\Big(\pi_\theta(y_w|x)\Big) \Big] \\
    & = \mathcal{L}(\pi_\theta, U) - \mathbb{E}_{(x,y_w) \sim \mathcal{D}}\Big [\log\Big(\pi_\theta(y_w|x)\Big) \Big].
\end{align*}
This results in the final formulation of our CPO loss function.

\section{Details And Influence of Human-Labeled Preference Data}
\label{app:sec:human_data}
\textit{TL;DR: Our analysis indicates that our human-labeled data has a relatively minimal impact, probably due to a high proportion of tied translations and potential human bias in the evaluation process.}

\subsection{Data Construction Details}
The human-labeled dataset we used is pair-wise and differs from the triplet format of our main dataset. It focuses exclusively on two language directions, \texttt{en}$\rightarrow$\texttt{de} and \texttt{en}$\rightarrow$\texttt{zh}, resulting in an additional 2K sentences. The English source sentences, selected from Wikipedia, undergo a filtering process to remove time stamps and URLs. Each sentence is translated using Google Translate and GPT-4, with human evaluators then assigning their preference between these two translations. The distribution of preferences, indicating the number of times translations from Google or GPT-4 were favored or instances where they tied, is detailed in Table \ref{app:tab:human_data}.

\begin{table}[ht]
\caption{The statistic of how many translations win or tie by each system evaluated by human. }
\vskip 0.15in
\label{app:tab:human_data}
\centering

\resizebox{0.45\linewidth}{!}{
\begin{tabular}{lccc}
\hline
      & Google Wins & GPT-4 Wins & Ties \\ \hline
\texttt{en}$\rightarrow$\texttt{de} & 418         & 435        & 203  \\
\texttt{en}$\rightarrow$\texttt{zh} & 362         & 412        & 282  \\ \hline
\end{tabular}
}
\vskip -0.1in
\end{table}

\subsection{Influence on Performance}
Given that our model operates in a many-to-many translation format and the additional data is specific to only \texttt{de} and \texttt{zh} directions, we anticipate changes in performance when translating into these languages, but not in others. To assess the impact of the human-labeled data, we conducted a comparison between models exclusively fine-tuned on triplet data and those fine-tuned on both triplet and human-labeled data. The training approach remained consistent, utilizing the ALMA-13B-LoRA model fine-tuned via CPO. It's important to note that tied data were excluded from this analysis due to their lack of clear preference.

\noindent\textbf{Results and Analysis}
We show the detailed results for \texttt{en}$\rightarrow$\texttt{xx} and \texttt{xx}$\rightarrow$\texttt{en} in Table \ref{app:tab:human_en_xx} and \ref{app:tab:human_xx_en}, respectively. The inclusion of human-labeled preference data does not significantly enhance overall translation performance. For \texttt{en}$\rightarrow$\texttt{zh}, marginal improvements are observed, though they are minimal. Conversely, for \texttt{en}$\rightarrow$\texttt{de}, a slight decline in performance is noted. In summary, the addition of human-labeled data shows no substantial difference in the \texttt{en}$\rightarrow$\texttt{xx} direction, and a minor decrease in performance for \texttt{xx}$\rightarrow$\texttt{en} on average. We hypothesize that the limited impact of these human-labeled data may stem from a high proportion of tied evaluations and potential human bias in the evaluation process. For instance, there are instances where the author consider GPT-4's translations to be superior, while human evaluators favor those produced by Google Translate.

\begin{table}[H]
\caption{A comparison of translation performance when utilizing solely triplet data versus a combination of triplet data and human-labeled data (our original setup) in the \texttt{en}$\rightarrow$\texttt{xx} direction. The \textbf{bold} number indicates superior performance. There is not obvious performance difference adding our human-labeled data.
}
\vskip 0.15in
\label{app:tab:human_en_xx}
\centering
\resizebox{1\linewidth}{!}{
\begin{tabular}{lccccccccc}
\hline
  \multirow{2}{*}{Dataset}                                & \multicolumn{3}{c}{\texttt{de}}                           & \multicolumn{3}{c}{\texttt{cs}}                           & \multicolumn{3}{c}{\texttt{is}}                           \\
                                  \cmidrule(lr){2-4} \cmidrule(lr){5-7} \cmidrule(lr){8-10} 
                      & KIWI-22        & KIWI-XXL       & XCOMET         & KIWI-22        & KIWI-XXL       & XCOMET         & KIWI-22        & KIWI-XXL       & XCOMET         \\ \hline
Only Triplet Data                 & \textbf{83.43} & \textbf{84.63} & \textbf{97.56} & 84.97          & \textbf{87.24} & 93.50          & 82.05          & 85.37          & 91.83          \\
Triplet Data + Human-Labeled Data & 83.28          & 84.25          & 97.48          & \textbf{84.99} & 87.06          & \textbf{93.61} & \textbf{82.18} & \textbf{85.68} & \textbf{91.93} \\ \hline
            \multirow{2}{*}{Dataset}                      & \multicolumn{3}{c}{\texttt{zh}}                           & \multicolumn{3}{c}{\texttt{ru}}                           & \multicolumn{3}{c}{Avg.}                         \\
                                  \cmidrule(lr){2-4} \cmidrule(lr){5-7} \cmidrule(lr){8-10} 
                        & KIWI-22        & KIWI-XXL       & XCOMET         & KIWI-22        & KIWI-XXL       & XCOMET         & KIWI-22        & KIWI-XXL       & XCOMET         \\ \hline
Only Triplet Data                 & 82.15          & 84.08          & 91.59          & \textbf{84.05} & \textbf{87.43} & \textbf{95.26} & 83.33          & \textbf{85.75} & 93.95          \\
Triplet Data + Human-Labeled Data & \textbf{82.25} & \textbf{84.32} & \textbf{92.03} & 83.98          & 87.37          & 95.22          & \textbf{83.34} & 85.74          & \textbf{94.05} \\ \hline
\end{tabular}
}
\vskip -0.1in
\end{table}

\begin{table}[H]
\caption{A comparison of translation performance when utilizing solely triplet data versus a combination of triplet data and human-labeled data (our original setup) in the \texttt{en}$\rightarrow$\texttt{xx} direction. The \textbf{bold} number indicates superior performance. Interestingly, the inclusion of our human-labeled data results in a slight decrease in average performance.}
\vskip 0.15in
\label{app:tab:human_xx_en}
\centering
\resizebox{1\linewidth}{!}{
\begin{tabular}{lccccccccc}
\hline
     \multirow{2}{*}{Dataset}                             & \multicolumn{3}{c}{\texttt{de}}                           & \multicolumn{3}{c}{\texttt{cs}}                           & \multicolumn{3}{c}{is}                           \\
                                  \cmidrule(lr){2-4} \cmidrule(lr){5-7} \cmidrule(lr){8-10} 
                    & KIWI-22        & KIWI-XXL       & XCOMET         & KIWI-22        & KIWI-XXL       & XCOMET         & KIWI-22        & KIWI-XXL       & XCOMET         \\ \hline
Only Triplet Data                 & \textbf{81.57} & \textbf{84.25} & \textbf{94.32} & \textbf{82.68} & 83.70          & 87.97          & \textbf{81.63} & \textbf{85.87} & 80.89          \\
Triplet Data + Human-Labeled Data & 81.50          & 83.97          & 94.20          & 82.63          & \textbf{83.75} & \textbf{88.03} & 81.57          & 85.73          & \textbf{80.49} \\ \hline
               \multirow{2}{*}{Dataset}                   & \multicolumn{3}{c}{\texttt{zh}}                           & \multicolumn{3}{c}{\texttt{ru}}                           & \multicolumn{3}{c}{Avg.}                         \\
                                  \cmidrule(lr){2-4} \cmidrule(lr){5-7} \cmidrule(lr){8-10} 
                        & KIWI-22        & KIWI-XXL       & XCOMET         & KIWI-22        & KIWI-XXL       & XCOMET         & KIWI-22        & KIWI-XXL       & XCOMET         \\ \hline
Only Triplet Data                 & \textbf{79.34} & \textbf{77.31} & \textbf{91.76} & \textbf{81.76} & \textbf{81.63} & \textbf{91.34} & \textbf{81.40} & \textbf{82.55} & \textbf{89.26} \\
Triplet Data + Human-Labeled Data & 79.24          & 77.17          & 91.65          & 81.72          & 81.54          & 91.18          & 81.33          & 82.43          & 89.11          \\ \hline
\end{tabular}
}
\vskip -0.1in
\end{table}

\section{WMT Winner Systems}
\label{app:sec:wmt_winner}
\subsection{Systems For WMT'21 And WMT'22}
The WMT competition winners for each direction as reported in WMT'21 and WMT'22 correspond to those used by \citet{gptmt}. For more detailed information, we direct readers to this paper.

\subsection{Systems For WMT'23}
For the \texttt{de}$\leftrightarrow$\texttt{en} and \texttt{zh}$\leftrightarrow$\texttt{en} language pairs, we selected the translation systems that attained the highest human rankings based on source-based Direct Assessment and Scalar Quality Metrics (DA+SQM). For \texttt{de}$\leftrightarrow$\texttt{ru}, in the absence of human rankings for these directions in \citet{wmt23}, we opted for the model with the highest COMET-22 scores as reported in \citet{wmt23}. Details about these models are available in Table \ref{app:tab:wmt23-winners}.
\begin{table}[ht]
\caption{The list of WMT'23 winners served for each language direction.}
\vskip 0.15in
\label{app:tab:wmt23-winners}
\centering
\resizebox{0.45\linewidth}{!}{
\begin{tabular}{lc}
\hline
Systems      & Language Pair \\ \hline
ONLINE-B     & en-de         \\
ONLINE-A     & de-en         \\
Lan-BridgeMT \citep{wu-hu-2023-exploring}  & en-zh         \\
Lan-BridgeMT \citep{wu-hu-2023-exploring} & zh-en         \\
ONLINE-G     & en-ru         \\
ONLINE-Y     & ru-en         \\ \hline
\end{tabular}
}
\vskip -0.1in
\end{table}

\section{Estimated Accuracy with Human Agreements}
\label{app:sec:estimated_acc}
In the paper, we adopt a new approach for highlighting improvements within tables, moving beyond the standard practice of specifying a static improvement threshold in metric $y$ by score $x$. Instead, our threshold is dynamic, calibrated to the minimal metric difference $x$ in metric $y$ that yields a perceptible distinction between two systems as recognized by humans \citep{kocmi2024navigating}. For instance, to align with human judgments at an 80\% concordance rate, the required improvement margin is $\geq 1.24$ for both KIWI-XXL and COMET-XXL, and $\geq 0.53$ for KIWI-22. A comprehensive delineation of these thresholds can be found in Table \ref{app:tab:estimated_acc}.

\begin{table}[H]
\caption{Thresholds and estimated accuracies for each metric used in our paper.}
\vskip 0.15in
\label{app:tab:estimated_acc}
\centering
\resizebox{0.85\linewidth}{!}{
\begin{tabular}{lcccccccccc}
\toprule
\bf \makecell[l]{Estimated \\ Accuracy} & \bf \makecell[c]{{\small Coin toss}\\ 50\%} & \bf 55\% & \bf 60\% & \bf 65\% & \bf 70\% & \bf 75\% & \bf 80\% & \bf 85\% & \bf 90\% & \bf 95\% \\
\midrule
BLEU                                 &                                        0.27 &     0.52 &     0.78 &     1.06 &     1.39 &     1.79 &     2.34 &     3.35 &        - &        - \\
Comet-22                  &                                        0.03 &     0.10 &     0.18 &     0.26 &     0.35 &     0.45 &     0.56 &     0.71 &     0.94 &     1.53 \\
KIWI-22  &                                        0.01 &     0.08 &     0.16 &     0.24 &     0.33 &     0.42 &     0.53 &     0.67 &     0.85 &     1.18 \\
XCOMET-XXL                &                                        0.02 &     0.19 &     0.37 &     0.56 &     0.76 &     0.98 &     1.24 &     1.55 &     1.99 &     2.74 \\
KIWI-XXL &                                        0.06 &     0.22 &     0.39 &     0.57 &     0.77 &     0.98 &     1.24 &     1.58 &     2.08 &     3.39 \\
\bottomrule
\end{tabular}
}
\vskip -0.1in
\end{table}

\section{Full Results of WMT'23}
\label{app:sec:wmt23_results}
The comprehensive results of WMT'23 are presented in Table \ref{app:tab:full_wmt23}. Similar to its performance in WMT'21 and WMT'22, ALMA-13B-R performs best on average among the SoTA translation models.
\begin{table}[H]
\caption{The full results of WMT'23. The highest score among all systems are bold. \compactbest{dark blue boxes} indicates that the improvement over the original ALMA model achieves \textit{at least} 80\% estimated accuracy with the human judgement \citep{kocmi2024navigating}, while the lesser improvements are highlighted in \compactbetter{shallow blue boxes}.}
\vskip 0.15in
\label{app:tab:full_wmt23}
\centering

\resizebox{1\linewidth}{!}{
\begin{tabular}{lccccccccc}
\hline
                         & \multicolumn{3}{c}{\texttt{de}$\rightarrow$\texttt{en}}                        & \multicolumn{3}{c}{\texttt{zh}$\rightarrow$\texttt{en}}                        & \multicolumn{3}{c}{\texttt{ru}$\rightarrow$\texttt{en}}                        \\
                         \cmidrule(lr){2-4} \cmidrule(lr){5-7} \cmidrule(lr){8-10}
                         & KIWI-22        & KIWI-XXL       & XCOMET         & KIWI-22        & KIWI-XXL       & XCOMET         & KIWI-22        & KIWI-XXL       & XCOMET         \\ \hline
Gold Reference           & 78.93          & 75.96          & 84.23          & 74.46          & 68.80          & 83.51          & 79.46          & 77.84          & 83.60          \\
WMT Winners              & 79.37          & 76.18          & 84.35          & \textbf{80.17} & \textbf{79.53} & 92.25          & 80.88          & 79.21          & 86.22          \\
TowerInstruct            & 79.67          & 77.60          & 86.28          & 79.84          & 78.13          & 91.75          & 80.85          & 80.03          & 87.76          \\
MADLAD-10B                                                                                                   & 78.52                           & 75.50                           & 83.85                           & 77.68                           & 73.72                           & 88.07                           & 79.65                           & 77.58                           & 85.15                           \\
ALMA-13B-LoRA            & 79.36          & 76.79          & 85.07          & 78.83          & 76.71          & 90.73          & 80.79          & 80.14          & 86.94          \\ \hdashline
+ CPO (Ours, ALMA-13B-R) & \better  \textbf{79.87} & \better \textbf{77.69} & \best 
 \textbf{86.62} & \best 80.01 & \best  78.42 & \best 
 \textbf{92.36} & \better  \textbf{81.11} & \better 
 \textbf{80.95} & \best 
 \textbf{88.75} \\ \hline
                         & \multicolumn{3}{c}{\texttt{en}$\rightarrow$\texttt{de}}                        & \multicolumn{3}{c}{\texttt{en}$\rightarrow$\texttt{zh}}                        & \multicolumn{3}{c}{\texttt{en}$\rightarrow$\texttt{ru}}                        \\
                         \cmidrule(lr){2-4} \cmidrule(lr){5-7} \cmidrule(lr){8-10}
                         & KIWI-22        & KIWI-XXL       & XCOMET         & KIWI-22        & KIWI-XXL       & XCOMET         & KIWI-22        & KIWI-XXL       & XCOMET         \\ \hline
Gold Reference           & 80.12          & \textbf{77.93} & 88.91          & 79.60          & 73.47          & 86.15          & 79.87          & 79.36          & 91.41          \\
WMT Winners              & \textbf{80.80} & 77.26          & 87.94          & 79.70          & 74.20          & 87.24          & \textbf{82.51} & 79.95          & 91.41          \\
TowerInstruct            & 80.13          & 75.34          & 86.55          & 80.03          & 74.85          & 86.74          & 81.33          & 77.14          & 89.59          \\
MADLAD-10B                                                                                                   & 77.48                           & 70.87                           & 86.18                           & 74.63                           & 62.07                           & 79.12                           & 79.24                           & 72.40                           & 86.64                           \\
ALMA-13B-LoRA            & 78.79          & 73.40          & 85.61          & 78.92          & 72.95          & 85.13          & 80.21          & 76.02          & 89.48          \\ \hdashline
+ CPO (Ours, ALMA-13B-R) & \best 79.85 &  \best 77.05 &  \best 
 \textbf{89.79} & \best \textbf{80.48} & \best \textbf{78.17} & \best  \textbf{88.34} & \best 81.97 &  \best \textbf{81.52} & \best 
 \textbf{92.56} \\ \hline
\end{tabular}
}
\vskip -0.1in
\end{table}

\section{Evaluation on ALMA-R with Non-Comet Metric}
\label{app:sec:bleurt}
Concerns may arise regarding the similar training procedure of COMET metrics, leading to high correlation among COMET models, which potentially undermine the validity of our analysis in Section \ref{sec:metric_preferred}. To address this, we also consider BLEURT-20 \citep{bleurt}, a non-COMET and neural-based (but reference-based evaluation) metric. We present BLEURT scores for ALMA-13B-LoRA and ALMA-13B-R in Table \ref{app:tab:bleurt}. Notably, even when preference data is constructed using COMET-based evaluations, significant improvements in non-COMET scores are observed. This strengthens our findings that translations produced by ALMA-R are indeed superior and robust.

\begin{table}[H]
\caption{The BLEURT-20 score comparsion between ALMA-13B-LoRA and ALMA-13B-R}
\vskip 0.15in
\label{app:tab:bleurt}
\centering

\resizebox{0.6\linewidth}{!}{

\begin{tabular}{lcccccc}
\hline
BLEURT-20        & \texttt{de}             & \texttt{cs}                & \texttt{is}                & \texttt{zh}                & \texttt{ru}                & Avg.           \\ \hline
\multicolumn{7}{c}{\textit{Translating to English} (\texttt{xx}$\rightarrow$\texttt{en})}                                                      \\
ALMA-13B-LoRA & 73.20          & 76.65          & 75.87          & 67.37          & 76.7           & 73.96          \\
ALMA-13B-R    & \textbf{73.62} & \textbf{76.94} & \textbf{76.98} & \textbf{69.48} & \textbf{76.91} & \textbf{74.79} \\ \hline
\multicolumn{7}{c}{\textit{Translating from English} (\texttt{en}$\rightarrow$\texttt{xx})}                                                    \\
ALMA-13B-LoRA & 75.51          & 80.93          & 73.19          & 70.54          & 74.94          & 75.02          \\
ALMA-13B-R    & \textbf{77.20} & \textbf{81.87} & \textbf{73.43} & \textbf{71.51} & \textbf{76.19} & \textbf{76.04} \\\hline

\end{tabular}

}
\vskip -0.1in
\end{table}

\section{The Effectiveness of The BC Regularizer for DPO}
\label{app:sec:regularizer_dpo}
The DPO loss $\mathcal{L}_\text{DPO} = \mathcal{L}(\pi_\theta, \pi_\text{ref})$ can also be utilized by adding our additional BC regularizer:
\begin{align*}
    \min_\theta \mathcal{L}(\pi_\theta, \pi_\text{ref}) - \mathbb{E}_{(x,y_w) \sim \mathcal{D}}\Big [\log\Big(\pi_\theta(y_w|x)\Big) \Big].
\end{align*}

In Table \ref{app:tab:dpo_nll}, we demonstrate that incorporating $\mathcal{L}_\text{NLL}$ into the DPO objective results in notable enhancements for translations both to and from English. 
This observation hints at why $\mathcal{L}_{\text{prefer}}$, as an approximation of $\mathcal{L}_\text{DPO}$, performs effectively, while the original DPO loss does not. It appears that the DPO loss lacks the BC regularizer, which steers the model towards the preferred data distribution. Although combining DPO with the BC regularizer could yield similar performance to CPO, it incurs double the memory cost and FLOPs per token in the forward pass. The original DPO loss shows the possibility of failure to improve the model performance in preference learning, so we here highlight the significance of incorporating BC regularization. Importantly, Table \ref{app:tab:dpo_nll} shows that $\mathcal{L}_{\text{prefer}}$ is a successful approximation of the DPO loss, offering savings in memory and speed, and it can even outperform the original BC-regularized DPO loss $\mathcal{L}_\text{DPO} + \mathcal{L}_\text{NLL}$.

\begin{table}[H]
\caption{The impact of applying $\mathcal{L}_\text{NLL}$ to the original DPO loss.}
\vskip 0.15in
\label{app:tab:dpo_nll}

\centering

\resizebox{0.7\linewidth}{!}{
\begin{tabular}{lccccc}
\hline
Loss Objective & KIWI-22        & KIWI-XXL       & XCOMET          & Memory Cost&FLOPs/tok\\ \hline
\multicolumn{6}{c}{\textit{Translating to English} (\texttt{xx}$\rightarrow$\texttt{en})}\\
$\mathcal{L}_\text{DPO}$& 80.51          & 81.36& 86.58           & 2$\times$&2$\times$
\\
$\mathcal{L}_\text{DPO}+\mathcal{L}_\text{NLL}$& 81.28& 82.42& 89.05 & 2$\times$&2$\times$
\\
$\mathcal{L}_\text{prefer}+\mathcal{L}_\text{NLL}$ (CPO)& \textbf{81.33} & \textbf{82.43} & \textbf{89.11}  & 1$\times$&1$\times$\\ \hline
\multicolumn{6}{c}{\textit{Translating from English} (\texttt{en}$\rightarrow$\texttt{xx})}\\
$\mathcal{L}_\text{DPO}$
& 82.27& 82.07& 92.25 & 2$\times$
&2$\times$
\\
$\mathcal{L}_\text{DPO}+\mathcal{L}_\text{NLL}$
& 83.13& 84.74& 93.53 & 2$\times$
&2$\times$
\\
$\mathcal{L}_\text{prefer}+\mathcal{L}_\text{NLL}$ (CPO) & \textbf{83.34} & \textbf{85.74} & \textbf{94.05}  & 1$\times$&1$\times$\\ \hline
\end{tabular}
}
\vskip -0.1in
\end{table}

\end{document}